\documentclass{article}
\usepackage[final]{colm2026_conference}

\usepackage{microtype}
\usepackage{hyperref}
\usepackage{url}
\usepackage{booktabs}
\usepackage[table]{xcolor}
\usepackage{enumitem}
\usepackage{graphicx}
\usepackage{multirow}
\usepackage{arydshln}
\usepackage{wrapfig}
\usepackage{amsmath}
\usepackage{nicematrix}
\usepackage{lineno}

\definecolor{darkblue}{rgb}{0, 0, 0.5}
\hypersetup{colorlinks=true, citecolor=darkblue, linkcolor=darkblue, urlcolor=darkblue}

\title{BERT-as-a-Judge: A Robust Alternative to Lexical Methods for Efficient Reference-Based LLM Evaluation}

\author{Hippolyte Gisserot-Boukhlef\textsuperscript{1,4} \\
Nicolas Boizard\textsuperscript{2,4} \quad Emmanuel Malherbe\textsuperscript{1} \quad 
\textmd{Céline Hudelot\textsuperscript{4} \quad
Pierre Colombo\textsuperscript{3}} \\[0.5em]
\textsuperscript{1}Artefact Research Center \quad
\textsuperscript{2}Diabolocom \quad
\textsuperscript{3}Cohere \\
\textsuperscript{4}MICS, CentraleSupélec, Université Paris-Saclay
}

\begin{document}

\ifcolmsubmission
\linenumbers
\fi

\maketitle

\begin{abstract}
Accurate evaluation is central to the large language model (LLM) ecosystem, guiding model selection and downstream adoption across diverse use cases. In practice, however, evaluating generative outputs typically relies on rigid lexical methods to extract and assess answers, which can conflate a model’s true problem-solving ability with its compliance with predefined formatting guidelines. While recent LLM-as-a-Judge approaches mitigate this issue by assessing semantic correctness rather than strict structural conformity, they also introduce substantial computational overhead, making evaluation costly. In this work, we first systematically investigate the limitations of lexical evaluation through a large-scale empirical study spanning 36 models and 15 downstream tasks, demonstrating that such methods correlate poorly with human judgments. To address this limitation, we introduce BERT-as-a-Judge, an encoder-driven approach for assessing answer correctness in reference-based generative settings, robust to variations in output phrasing, and requiring only lightweight training on synthetically annotated question-candidate-reference triplets. We show that it consistently outperforms the lexical baseline while matching the performance of much larger LLM judges, providing a compelling trade-off between the two and enabling reliable, scalable evaluation. Finally, through extensive experimentation, we provide detailed insights into BERT-as-a-Judge’s performance to offer practical guidance for practitioners, and release all project artifacts to foster downstream adoption.
\end{abstract}

\section{Introduction}

Evaluation lies at the core of the large language model (LLM) ecosystem. In recent years, considerable effort has been devoted to rigorously and fairly assessing model performance across a wide range of tasks, to guide model selection and downstream adoption \citep{liang2022helm,bommasani2021foundation}. For instruction-tuned models (optimized for human interaction and question answering), evaluation is typically conducted in zero-shot generative settings \citep{wei2022finetuned,ouyang2022training}, in which models are prompted to directly generate an answer without access to task-specific examples. 

While conceptually straightforward, this setup poses two challenges for evaluation: reliably extracting the model’s predicted answer for comparison with a reference, and performing the comparison itself. The former arises from answer formatting variations, such as ``The answer is X'' versus ``Answer: X'', whereas the latter occurs when comparing outputs like ``2.00'' versus ``2\$'', both of which should be treated as equivalent. A common mitigation strategy is to enforce constrained output formats via prompting, enabling answers to be extracted with regular expressions (regex) \citep{liang2023holistic, eval-harness}, and then rely on metrics beyond exact match, such as ROUGE \citep{rouge}, BERTScore \citep{bertscore}, or Math-Verify \citep{mathverify}, thereby avoiding errors caused by formatting inconsistencies or lexical variations. Although more flexible than strict exact match, these metrics can still fail to accurately capture answer correctness, especially since models often do not strictly follow prescribed output formats, making reliable answer parsing difficult. Such deviations may stem from differences in model scale, instruction-tuning data mixtures, or alignment strategies, and can artificially deflate measured downstream performance. While formatting adherence is itself an important capability, particularly for instruction following and structured generation \citep{ouyang2022training}, it should not confound the evaluation of orthogonal competencies such as factual knowledge, mathematical reasoning, or reading comprehension \citep{mmlu,gsm8k}.

Recently, LLM-as-a-Judge frameworks have emerged as a compelling alternative \citep{zheng2023judging,wang2023judgelm}. By delegating answer comparison to a separate language model, these approaches reduce dependence on rigid formatting constraints and can correctly credit semantically valid but structurally unconventional responses. However, they introduce substantial computational overhead and additional sources of variance, including sensitivity to the choice of judge model and prompt design \citep{whendoesreasoning}.

\begin{center}
\fbox{\parbox{0.95\textwidth}{%
\textbf{Question.} How can we measure a model’s core problem-solving ability without relying on output formatting or expensive inference?
}}
\end{center}

\begin{figure}[t]
    \centering
    \includegraphics[width=0.925\textwidth]{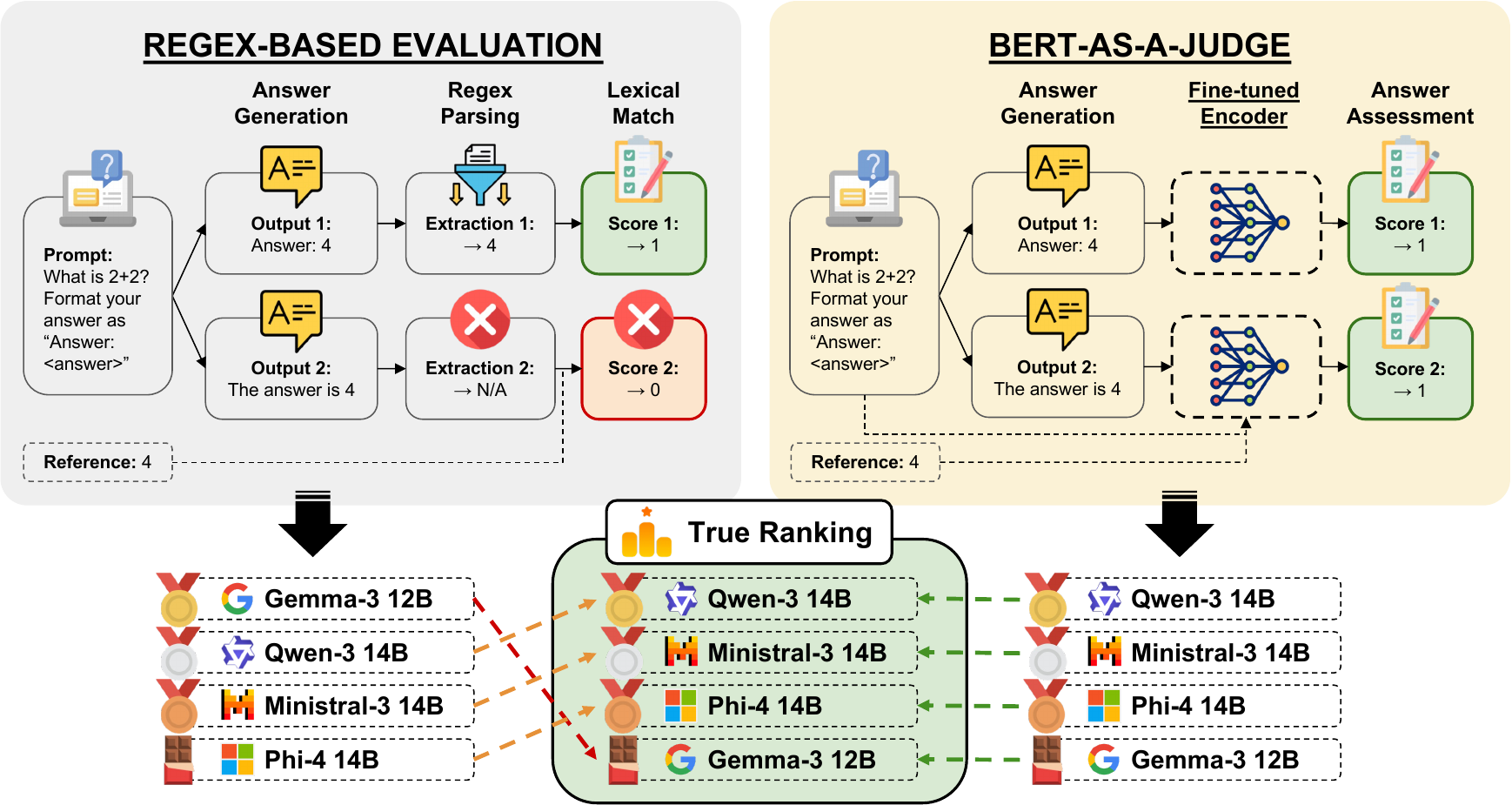}
    \caption{Comparison between regex-based (lexical) evaluation and BERT-as-a-Judge. Top: illustration of both approaches with simple examples. Bottom: model rankings for four similarly sized models from different families, computed via task-wise Borda count.}
    \label{fig:figure_1}
\end{figure}

\paragraph{Contributions.} In this work, we make the following three contributions:

\begin{itemize}[leftmargin=15pt, itemsep=1pt, topsep=0pt]    
    \item Through a comprehensive empirical study across a diverse set of models and tasks, we show that lexical evaluation exhibits weak correlation with human judgments (\autoref{sec:regex_limitations}).

    \item To address this limitation, we introduce BERT-as-a-Judge, an encoder-driven approach for evaluating generative models in reference-based settings, leveraging the strength of bidirectional attention for text classification (\autoref{fig:figure_1}). We show that BERT-as-a-Judge consistently outperforms lexical evaluation and even surpasses LLM-as-a-Judge under comparable inference conditions (\autoref{sec:encoder_based_evaluation}).

    \item We provide detailed insights into BERT-as-a-Judge’s performance through an extensive set of experiments, offering practical guidance for downstream applications (\autoref{sec:xp_analysis}). Additionally, we release the packaged code\footnote{\url{https://github.com/artefactory/Bert-as-a-Judge}} along with the full set of generated and annotated data, covering outputs from 36 models across 15 tasks, and open-source all fine-tuned checkpoints used in our experiments.\footnote{\url{https://hf.co/collections/artefactory/bert-as-a-judge}}
\end{itemize}

\section{Experimental Protocol}

\subsection{Answer Generation}
\label{sec:dataset_generation}

\paragraph{Tasks.}

The backbone of LLM evaluation consists of tasks whose outputs can be unambiguously judged as correct or incorrect, providing an objective basis for model assessment \citep{llama3,qwen3,olmo3,eurollm22b,apertus}. In this work, we focus on three families of widely used benchmarks:

\begin{itemize}[leftmargin=15pt, itemsep=1pt, topsep=0pt]
    \item \emph{Multiple-choice}, in which models are given a question along with a set of options: MMLU \citep{mmlu}, MMLU-Pro \citep{mmlupro}, TruthfulQA \citep{truthfulqa}, ARC-Easy/Challenge \citep{arc}, and GPQA \citep{gpqa}.
    \item \emph{Context extraction}, where models must provide answers grounded in a given passage by citing relevant evidence: SQuAD-v2 \citep{squadv2}, HotpotQA \citep{hotpotqa}, DROP \citep{drop}, and CoQA \citep{coqa}.
    \item \emph{Open-form mathematics}, in which models generate a final closed-form answer in free text: GSM8K \citep{gsm8k}, MATH \citep{math}, AsDiv \citep{asdiv}, AIME 24 \citep{aime24}, and AIME 25 \citep{aime25}.
\end{itemize}

\paragraph{Models.}
We perform inference across a broad range of recent open-weight instruction-tuned model families, spanning from 135M to 70B parameters. Our study includes 36 models in total: Llama-3 (1B, 3B, 8B, 70B) \citep{llama3}, Qwen-3 (600M, 4B, 8B, 14B, 32B) \citep{qwen3}, Gemma-3 (1B, 4B, 12B, 27B) \citep{gemma3}, Falcon-3 (1B, 3B, 7B) \citep{falcon3}, Phi-4 (3.8B, 14B) \citep{phi4,phi4mini}, SmolLM-2 and 3 (135M, 360M, 1.7B, 3B) \citep{smollm2,smollm3}, OLMo-3 (7B, 32B) \citep{olmo3}, Ministral-3 (3B, 8B, 14B) \citep{ministral3}, LFM-2 (350M, 700M, 1.2B, 2.6B) \citep{ministral3}, EuroLLM (1.7B, 9B, 22B) \citep{eurollm,eurollm9b,eurollm22b}, and Apertus (8B, 70B) \citep{apertus}.

\paragraph{Generation parameters.} 
For each task-model pair, responses are produced in a zero-shot setting using greedy decoding, with a maximum generation length of 2048 tokens. For experimental purposes, models are prompted to conclude their outputs in the format ``Final answer: [answer]'' to facilitate downstream regex parsing and ensure fair comparison between model- and regex-based assessment methods.


\subsection{Labeling}
\label{sec:labeling}

\paragraph{Synthetic labeling.}
For annotation, we employ Nemotron-Super-v1.5 \citep{nemotron} as an automatic evaluator. The model is provided with the question, the candidate answer, and the reference answer, and is asked to determine whether the candidate response is correct given the available information.\footnote{The full evaluation prompt is provided in \autoref{sec:prompting_details}.} Inference is conducted using greedy decoding in non-reasoning mode.

\paragraph{Human labeling.} To validate the reliability of the synthetic labeling approach, we perform human annotation on a subset of the data. Specifically, we randomly sample instances from the generated dataset and have them independently labeled by a pool of 11 human evaluators, totaling 3,212 annotations, and resulting in an overall average agreement of 97.5\% with the synthetic labels.\footnote{Further details are provided in \autoref{sec:human_labels}.}

\subsection{Evaluation Methods}
\label{sec:evaluation_methods}

\paragraph{BERT-as-a-Judge.} 
We propose to train a BERT-style encoder model on labeled question-candidate-reference triplets constructed as described in \autoref{sec:dataset_generation} and \autoref{sec:labeling}, leveraging its bidirectional attention mechanism, which is well suited for structured text classification \citep{zhang2025bert}. We construct the training mixture from the tasks described in \autoref{sec:dataset_generation} that provide an explicit training split, namely MMLU, ARC-Easy, ARC-Challenge, SQuAD-v2, HotpotQA, GSM8K, and Math. The training dataset is constructed to balance the number of samples across task categories and models, resulting in approximately 1M synthetically labeled samples in total. We initialize the encoder from EuroBERT 210M \citep{eurobert} and fine-tune it for one epoch using binary cross-entropy. We employ a learning rate of $2 \times 10^{-5}$, following the authors’ recommendations for sequence classification, along with a 5\% warmup ratio and a linear decay schedule. Training is conducted on 8 MI250x GPUs, yielding an effective batch size of 32, taking approximately 20 GPU hours per run.

\paragraph{Baselines.} 
We compare BERT-as-a-Judge to the following baselines:

\begin{itemize}[leftmargin=15pt, itemsep=1pt, topsep=0pt]   
    \item \emph{Regex:} Extracts answers using a regular expression based on the pattern ``Final answer: [answer]'' and evaluates multiple-choice tasks with exact match, context extraction with ROUGE-L \citep{rouge}, and open-form math with Math-Verify \citep{mathverify}. For answer parsing, we build on the regex rules provided by the lm-evaluation-harness framework \citep{eval-harness}, adapting them to our prompting format and the range of evaluated models.
        
    \item \emph{LLM-as-a-Judge:} Uses a generative language model to determine whether a candidate response matches the reference answer for a given question, following the procedure described in \autoref{sec:dataset_generation}. We evaluate three categories of generative judges: (i)~off-the-shelf instruction-tuned models, including Qwen-3 (0.6B to 14B) and Gemma-3 (1B to 27B); (ii)~a fine-tuned version of Qwen-3 0.6B, trained on the same dataset as BERT-as-a-Judge under an equivalent training FLOP budget; and (iii)~the JudgeLM family (7B, 13B, and 33B) \citep{wang2023judgelm}, a collection of models specifically optimized for generative evaluation. While JudgeLM natively produces a score from 1 to 10,\footnote{A response is considered correct if the score exceeds 5.} Qwen-3 and Gemma-3 are prompted to output either ``True'' or ``False'' to indicate whether the candidate response is correct. We consider two prompting strategies: direct assessment (denoted ``S'' for ``short'') and assessment with an intermediate reasoning step before the final prediction (denoted ``L'' for ``long'').
        
    \item \emph{Encoder-based approach:} Relies on a neural encoder metric to assess answer correctness. Specifically, we evaluate BLEURT-base and BLEURT-large \citep{bleurt}.
\end{itemize}

\subsection{Assessment of Evaluation Quality}
\label{sec:assessment}

\paragraph{Metric.} We assess each method by its accuracy against synthetic labels from Nemotron-Super-v1.5,\footnote{For methods producing scores between 0 and 1 (e.g., encoder models with soft probabilities), we use a default threshold of 0.5.} reflecting how well it predicts whether a given answer is correct or incorrect.\footnote{\autoref{sec:human_labels} provides a detailed analysis of the agreement between synthetic and human labels, showing that evaluating on synthetic labels does not introduce any substantial bias.}

\paragraph{Benchmarks.} 
We assess all evaluation methods on the full set of tasks introduced in \autoref{sec:dataset_generation}, including both the test splits of tasks used during encoder training (\autoref{sec:evaluation_methods}) and the tasks reserved exclusively for out-of-domain evaluation.\footnote{We also provide preliminary multilingual results on the translated version of MMLU \citep{dac2023okapi} in \autoref{sec:multilingual_results}.}

\section{Limitations of Regex-Based Evaluation}
\label{sec:regex_limitations}

\begin{figure}[h]
    \centering
    \includegraphics[width=\textwidth]{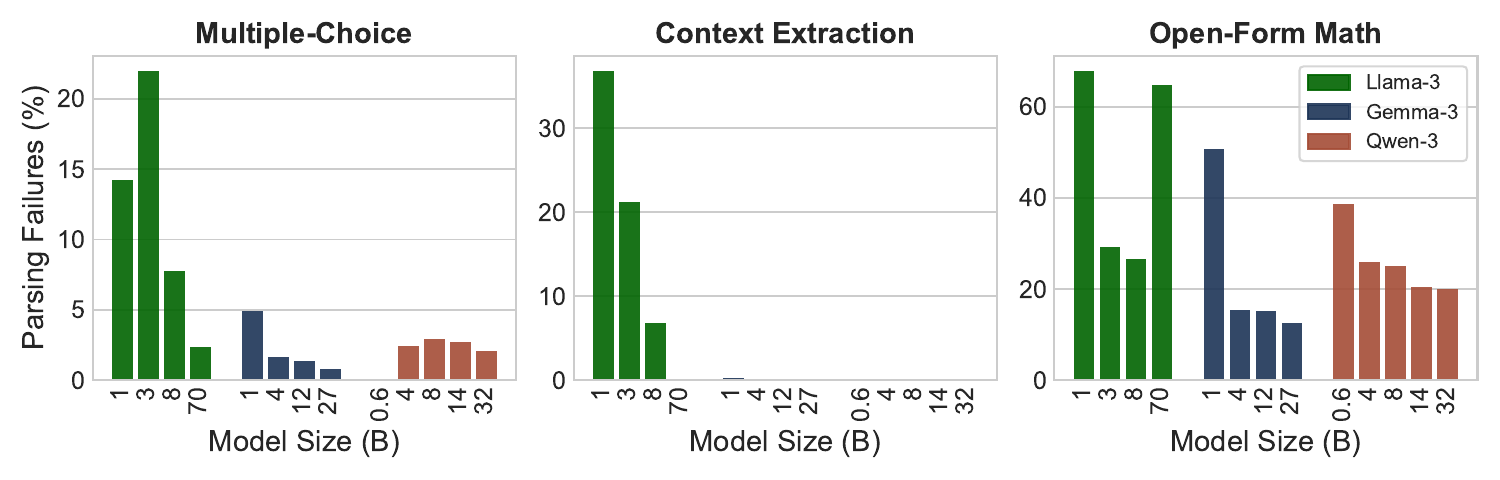}
    \vspace{-20pt}
    \caption{Quantification of regex parsing failures. Values represent the failure rate, defined as the percentage of instances with unparsable outputs. Results are shown for the Llama-3, Gemma-3, and Qwen-3 model families and aggregated by task category.}
    \label{fig:regex_failures}
\end{figure}

This section analyzes the impact of regex-based evaluation on measured downstream performance, noting discrepancies from both formatting-related parsing failures and post-parsing matching errors. Specifically, we quantify parsing failure rates across a range of models in \autoref{fig:regex_failures} and assess performance deltas relative to ground-truth labels in \autoref{tab:regex_impact}.

\paragraph{Model scale, family, and task type have a high impact on output formatting.}
\autoref{fig:regex_failures} shows that larger models tend to produce fewer formatting errors, as illustrated by the Llama-3 models on context extraction and Qwen-3 on open-form math tasks. Model family also plays a significant role: Qwen-3 and Gemma-3 consistently achieve near-perfect formatting compliance on context extraction, whereas smaller Llama-3 models exhibit substantially higher failure rates. Task type further impacts formatting accuracy. Open-form math proves the most challenging, with Llama-3 70B generating incorrectly formatted outputs over 60\% of the time and Qwen-3 32B around 20\%, while multiple-choice and context extraction tasks are much easier, with mid- to large-scale models often achieving near-zero failure rates.

\begin{table}[h]
    \centering
    \small
    \begin{tabular}{llcccccc}
\toprule
\multirow{2}{*}{\textbf{Family}} & \multirow{2}{*}{\textbf{Size}} & \multicolumn{2}{c}{\textbf{Multiple-Choice}} & \multicolumn{2}{c}{\textbf{Context Extraction}} & \multicolumn{2}{c}{\textbf{Open-Form Math}} \\
\cmidrule(lr){3-4} \cmidrule(lr){5-6} \cmidrule(lr){7-8}
 &  & \multicolumn{1}{c}{$\Delta$Accuracy} & \multicolumn{1}{c}{$\Delta$Rank} & \multicolumn{1}{c}{$\Delta$Accuracy} & \multicolumn{1}{c}{$\Delta$Rank} & \multicolumn{1}{c}{$\Delta$Accuracy} & \multicolumn{1}{c}{$\Delta$Rank} \\
\midrule
\multirow{4}{*}{Llama-3} 
 & 1B & -0.9 {\tiny (-2.6/+1.7)} & \cellcolor{green!44}$\uparrow$4.8 & -23.9 {\tiny (-18.1/-5.8)} & \cellcolor{red!13}$\downarrow$1.2 & -11.9 {\tiny (-11.1/-0.8)} & \cellcolor{red!12}$\downarrow$0.8 \\
 & 3B & -13.1 {\tiny (-8.5/-4.6)} & \cellcolor{red!11}$\downarrow$0.5 & -27.5 {\tiny (-15.6/-11.9)} & \cellcolor{red!11}$\downarrow$0.4 & -7.0 {\tiny (-2.6/-4.4)} & \cellcolor{green!22}$\uparrow$1.7 \\
 & 8B & -23.3 {\tiny (-0.9/-22.3)} & \cellcolor{red!28}$\downarrow$6.4 & -21.0 {\tiny (-4.0/-17.0)} & \cellcolor{green!35}$\uparrow$3.5 & -5.4 {\tiny (-0.8/-4.6)} & \cellcolor{green!40}$\uparrow$4.2 \\
 & 70B & -1.1 {\tiny (-1.1/-0.1)} & \cellcolor{green!19}$\uparrow$1.3 & -18.4 {\tiny (-0.1/-18.3)} & \cellcolor{green!56}$\uparrow$6.5 & -30.4 {\tiny (-29.0/-1.4)} & \cellcolor{red!48}$\downarrow$13.3 \\
\midrule
\multirow{4}{*}{Gemma-3} 
 & 1B & -0.1 {\tiny (-0.7/+0.5)} & \cellcolor{green!51}$\uparrow$5.7 & -12.3 {\tiny (-0.1/-12.2)} & \cellcolor{green!43}$\uparrow$4.6 & -11.3 {\tiny (-6.8/-4.5)} & \cellcolor{red!13}$\downarrow$1.2 \\
 & 4B & +0.3 {\tiny (-0.2/+0.5)} & \cellcolor{green!54}$\uparrow$6.2 & -30.8 {\tiny (-0.0/-30.8)} & \cellcolor{red!14}$\downarrow$1.5 & -11.8 {\tiny (-1.2/-10.6)} & \cellcolor{red!10}$\downarrow$0.1 \\
 & 12B & -0.1 {\tiny (-0.2/+0.1)} & \cellcolor{green!48}$\uparrow$5.3 & -27.6 {\tiny (-0.1/-27.4)} & \cellcolor{red!22}$\downarrow$4.2 & -10.3 {\tiny (-1.4/-8.9)} & \cellcolor{green!21}$\uparrow$1.6 \\
 & 27B & +0.1 {\tiny (-0.1/+0.2)} & \cellcolor{green!44}$\uparrow$4.7 & -28.7 {\tiny (-0.0/-28.7)} & \cellcolor{red!16}$\downarrow$2.1 & -10.8 {\tiny (-1.1/-9.7)} & \cellcolor{green!13}$\uparrow$0.4 \\
\midrule
\multirow{5}{*}{Qwen-3} 
 & 0.6B & -20.8 {\tiny (-0.0/-20.8)} & \cellcolor{red!18}$\downarrow$2.8 & -13.8 {\tiny (-0.0/-13.8)} & \cellcolor{green!30}$\uparrow$2.8 & -10.8 {\tiny (-5.8/-5.0)} & \cellcolor{red!11}$\downarrow$0.5 \\
 & 4B & -3.2 {\tiny (-0.5/-2.7)} & \cellcolor{green!35}$\uparrow$3.5 & -20.2 {\tiny (-0.0/-20.2)} & \cellcolor{green!60}$\uparrow$7.0 & -12.0 {\tiny (-5.3/-6.7)} & \cellcolor{green!17}$\uparrow$1.0 \\
 & 8B & -7.0 {\tiny (-0.6/-6.3)} & \cellcolor{green!15}$\uparrow$0.7 & -29.7 {\tiny (-0.0/-29.7)} & \cellcolor{red!30}$\downarrow$7.2 & -15.2 {\tiny (-7.9/-7.3)} & \cellcolor{red!11}$\downarrow$0.2 \\
 & 14B & -19.3 {\tiny (-0.7/-18.6)} & \cellcolor{red!47}$\downarrow$13.2 & -20.9 {\tiny (-0.0/-20.9)} & \cellcolor{green!14}$\uparrow$0.5 & -15.0 {\tiny (-7.8/-7.2)} & \cellcolor{green!16}$\uparrow$0.8 \\
 & 32B & -23.8 {\tiny (-0.5/-23.3)} & \cellcolor{red!60}$\downarrow$17.6 & -25.1 {\tiny (-0.0/-25.1)} & \cellcolor{red!13}$\downarrow$1.1 & -10.3 {\tiny (-2.6/-7.7)} & \cellcolor{green!30}$\uparrow$2.8 \\
\bottomrule
\end{tabular}
    \caption{Impact of regex-based evaluation on measured model performance. The ``$\Delta$Accuracy'' columns report the difference between regex-based and ground-truth (synthetic-label) accuracy, with values in parentheses indicating the contributions from parsing failures and post-parsing matching, respectively. The ``$\Delta$Rank'' columns show the corresponding average change in ranking across the 36 evaluated models, where greener values indicate upward rank changes and redder values indicate downward rank changes.}
    \label{tab:regex_impact}
\end{table}

\paragraph{Regex-based evaluation distorts performance measurements.}
\autoref{tab:regex_impact} illustrates the risks of relying on regex-based evaluation, showing substantial negative deltas in measured performance across a broad set of models. Notably, even models with high formatting compliance (e.g., Gemma-3 family on context extraction tasks) suffer from substantial underestimation, often due to overly verbose outputs that technically follow formatting rules but fail lexical matching.\footnote{More details are provided in \autoref{sec:examples}.} Overall, regex-based assessment significantly distorts leaderboard rankings; for instance, Qwen-3 32B drops 18 positions while Gemma-3 4B climbs 6 places on multiple-choice tasks, with these shifts often representing mere artifacts of formatting quirks and rigid lexical matching rather than true differences in capability.

\section{Encoder-Based Evaluation}
\label{sec:encoder_based_evaluation}

\begin{figure}[h]
    \centering
    \includegraphics[width=\textwidth]{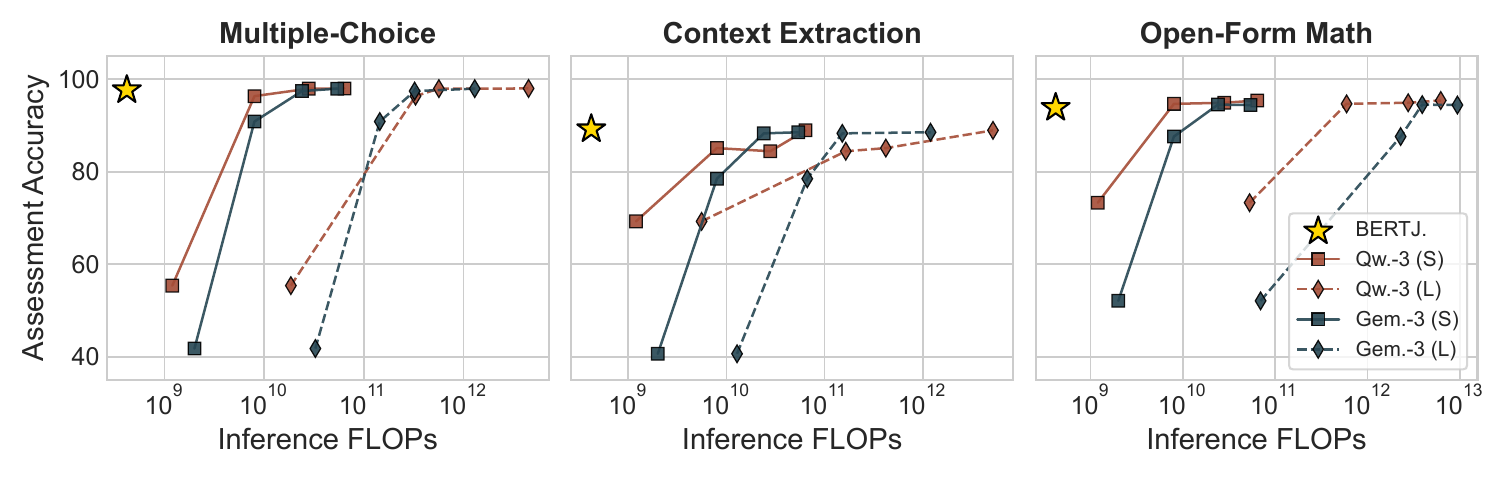}
    \vspace{-20pt}
    \caption{Compute-aware comparison between BERT-as-a-Judge and LLM judges from the Qwen-3 and Gemma-3 families under S and L prompting strategies.}
    \label{fig:generator_scaling}
\end{figure}

In this section, motivated by the observation that regex-based evaluation fails to accurately reflect true model performance across a wide range of benchmarks (\autoref{sec:regex_limitations}), we train a BERT-as-a-Judge encoder model to assess answer correctness following the methodology described in \autoref{sec:evaluation_methods}. We then assess the proposed approach using the setup detailed in \autoref{sec:assessment} and report the results in \autoref{tab:main_results}. To further support our analysis, \autoref{tab:model_removal} reports performance for models whose outputs were excluded from the training mixture, while \autoref{fig:generator_scaling} provides a compute-aware comparison with LLM-as-a-Judge methods.

\begin{table}[h]
\centering
\small
\resizebox{\textwidth}{!}{
\begin{tabular}{l *{11}{c}}
\toprule
\multirow{2}{*}{\textbf{Task}} & \multirow{2}{*}{\textbf{Regex}} & \multicolumn{3}{c}{\textbf{Qwen-3}} & \textbf{Gemma-3} & \multicolumn{3}{c}{\textbf{JudgeLM}} & \multicolumn{2}{c}{\textbf{BLEURT}} & \multirow{2}{*}{\textbf{BERTJudge}} \\
\cmidrule(lr){3-5} \cmidrule(lr){6-6} \cmidrule(lr){7-9} \cmidrule(lr){10-11}
& & \textbf{0.6B} & \textbf{0.6B (FT)} & \textbf{14B} & \textbf{12B} & \textbf{7B} & \textbf{13B} & \textbf{33B} & \textbf{Base} & \textbf{Large} & \\
\midrule
\multicolumn{12}{c}{\textit{Multiple-Choice}} \\
\addlinespace[2pt]
ARC-Challenge & 89.0 & 50.2 & 98.1 & 99.1 & \textbf{99.4} & 59.1 & 60.8 & 65.4 & 23.8 & 23.8 & \textbf{99.4} \\
ARC-Easy      & 88.2 & 54.0 & 99.1 & 98.9 & \textbf{99.7} & 53.2 & 53.8 & 59.5 & 15.1 & 15.1 & \textbf{99.7} \\
MMLU          & 88.1 & 50.2 & 94.3 & \textbf{98.6} & 98.3 & 57.4 & 60.6 & 66.6 & 37.9 & 38.0 & \textbf{98.5} \\
\addlinespace[0.5pt] \cdashline{1-12} \addlinespace[3pt]
GPQA          & 68.0 & 65.9 & 82.0 & \textbf{95.3} & 92.5 & 64.9 & 70.6 & 73.5 & 68.0 & 68.1 & \textbf{93.5} \\
MMLU-Pro      & 88.7 & 56.9 & 87.5 & \textbf{97.0} & 96.1 & 60.4 & 64.4 & 68.7 & 55.5 & 55.5 & \textbf{96.4} \\
TruthfulQA    & 92.5 & 54.5 & 94.8 & \textbf{98.9} & 98.5 & 63.2 & 61.3 & 65.5 & 48.4 & 48.4 & \textbf{98.6} \\
\midrule
\multicolumn{12}{c}{\textit{Context Extraction}} \\
\addlinespace[2pt]
HotpotQA      & 75.6 & 70.0 & \textbf{90.1} & 86.3 & 89.6 & 31.0 & 35.4 & 42.2 & 29.0 & 29.0 & \textbf{90.8} \\
SQuAD-v2      & 72.3 & 62.7 & \textbf{84.9} & 81.1 & 81.1 & 45.4 & 44.3 & 53.5 & 46.5 & 46.5 & \textbf{89.3} \\
\addlinespace[0.5pt] \cdashline{1-12} \addlinespace[3pt]
CoQA          & 67.0 & 75.2 & \textbf{89.8} & 80.9 & \textbf{90.0} & 26.3 & 29.5 & 31.8 & 23.5 & 23.5 & 88.2 \\
DROP          & 77.0 & 69.3 & \textbf{90.5} & 89.5 & \textbf{92.5} & 42.7 & 54.6 & 49.3 & 39.8 & 39.8 & 88.2 \\
\midrule
\multicolumn{12}{c}{\textit{Open-Form Math}} \\
\addlinespace[2pt]
GSM8K         & 94.4 & 71.2 & 97.2 & \textbf{99.0} & 98.6 & 26.7 & 43.5 & 44.3 & 26.4 & 26.4 & \textbf{98.8} \\
Math          & 73.3 & 58.7 & 90.5 & 93.6 & \textbf{94.1} & 39.9 & 44.0 & 45.5 & 39.6 & 39.5 & \textbf{93.7} \\
\addlinespace[0.5pt] \cdashline{1-12} \addlinespace[3pt]
AIME24        & 87.7 & 77.6 & 87.3 & \textbf{90.5} & \textbf{90.5} & 80.0 & 80.0 & 80.1 & 80.0 & 80.0 & 89.8 \\
AIME25        & 91.6 & 82.7 & 86.3 & \textbf{93.5} & \textbf{92.8} & 85.3 & 85.3 & 85.3 & 85.3 & 85.2 & 92.5 \\
ASDiv         & 89.2 & 75.5 & 95.9 & \textbf{97.8} & \textbf{96.3} & 18.9 & 31.0 & 51.2 & 18.4 & 18.4 & 95.1 \\
\bottomrule
\end{tabular}
}
\caption{Accuracy of evaluation methods against ground-truth labels across tasks, averaged over models. Dashed lines separate test-only tasks from tasks with an available training split. Qwen-3 and Gemma-3 models use direct assessment (S method). ``FT'' denotes the fine-tuned Qwen-3 0.6B model. Bold indicates the two highest accuracies for each task.}
\label{tab:main_results}
\end{table}

\paragraph{BERT-as-a-Judge shows the strongest overall alignment with human judgments.}
As shown in \autoref{tab:main_results}, our trained encoder consistently matches or surpasses the performance of off-the-shelf generative judges up to 70 times larger in size (Qwen-3 14B and Gemma-3 12B), while also outperforming a fine-tuned generative model with three times more parameters (Qwen-3 0.6B (FT)). This trend holds across all task categories, achieving near-perfect alignment on multiple-choice datasets (e.g., 99.7\% on ARC-Easy and 98.5\% on MMLU) while providing substantial gains on context extraction tasks. Furthermore, it significantly outperforms the regex baseline (e.g., +21.2\% on CoQA, +20.4\% on MATH, and +10.4\% on ARC-Challenge), whereas off-the-shelf Qwen-3 0.6B, JudgeLM, and BLEURT fail to generalize effectively across the considered tasks. These findings demonstrate that a dedicated, lightweight encoder can assess answer correctness in a highly reliable manner while maintaining high computational efficiency ($\approx$200 ms/sample on an Apple M1 CPU).

\paragraph{BERT-as-a-Judge is robust to out-of-domain tasks.}
Beyond its strong overall accuracy, the encoder-based method maintains high accuracy even on tasks excluded from the training mixture (e.g., 98.6\% on TruthfulQA, 88.1\% on CoQA, and 95.3\% on ASDiv), highlighting its strong generalization ability across the three task categories considered (\autoref{tab:main_results}).

\begin{wraptable}{r}{0.45\textwidth}
\vspace{3pt}
\centering
\small
\resizebox{0.45\textwidth}{!}{
\begin{tabular}{lcccccc}
\toprule
\multirow{3}{*}{\textbf{Size}} & \multicolumn{2}{c}{\textbf{Multiple-}} & \multicolumn{2}{c}{\textbf{Context}} & \multicolumn{2}{c}{\textbf{Open-Form}} \\
 & \multicolumn{2}{c}{\textbf{Choice}} & \multicolumn{2}{c}{\textbf{Extraction}} & \multicolumn{2}{c}{\textbf{Math}} \\
\cmidrule(lr){2-3} \cmidrule(lr){4-5} \cmidrule(lr){6-7}
& \textbf{ID} & \textbf{OOD} & \textbf{ID} & \textbf{OOD} & \textbf{ID} & \textbf{OOD} \\
\midrule
\multicolumn{7}{c}{\textit{Ministral-3}} \\
\addlinespace[2pt]
3B  & 97.0 & 96.9 & 83.7 & 82.5 & 81.4 & 86.5 \\
8B  & 97.9 & 97.8 & 87.9 & 87.1 & 83.5 & 84.8 \\
14B & 98.2 & 98.3 & 89.1 & 88.6 & 83.5 & 82.6 \\
\midrule
\multicolumn{7}{c}{\textit{LFM-2}} \\
\addlinespace[2pt]
0.35B & 94.8 & 94.1 & 90.5 & 88.6 & 97.1 & 96.9 \\
0.7B  & 96.8 & 96.7 & 87.4 & 84.4 & 97.4 & 97.0 \\
1.2B  & 97.5 & 97.1 & 91.2 & 90.9 & 94.6 & 94.4 \\
2.6B  & 97.9 & 97.8 & 87.1 & 86.0 & 93.9 & 93.7 \\
\midrule
\multicolumn{7}{c}{\textit{EuroLLM}} \\
\addlinespace[2pt]
1.7B & 93.4 & 93.1 & 91.5 & 91.2 & 98.5 & 98.4 \\
9B   & 98.6 & 98.5 & 90.7 & 90.2 & 94.5 & 94.1 \\
22B  & 98.2 & 98.1 & 90.6 & 90.6 & 91.5 & 91.1 \\
\midrule
\multicolumn{7}{c}{\textit{Apertus}} \\
\addlinespace[2pt]
8B  & 97.6 & 97.4 & 90.3 & 89.9 & 97.5 & 97.4 \\
70B & 98.1 & 98.0 & 89.5 & 89.5 & 97.2 & 97.1 \\
\bottomrule
\end{tabular}
}
\caption{Assessment accuracy on out-of-domain models. ``ID'' denotes training on all model outputs, while ``OOD'' excludes specific models from the training mixture. Results are aggregated by task category.}
\label{tab:model_removal}
\vspace{-10pt}
\end{wraptable}

\paragraph{BERT-as-a-Judge generalizes to unseen models.} 
\autoref{tab:model_removal} shows that removing generations from specific models in the training mixture has minimal impact on downstream assessment quality for those excluded instances. This demonstrates the strong generalization ability of our approach, indicating it can be safely extended to additional model families outside the training mixture and supporting broader adoption.

\paragraph{BERT-as-a-Judge achieves a Pareto-optimal accuracy-efficiency trade-off.}
To complement the findings of \autoref{tab:main_results}, we conduct an extensive comparison with LLM-as-a-Judge methods by evaluating the Qwen-3 and Gemma-3 families across a wide range of model sizes (0.6B to 32B for Qwen-3 and 1B to 27B for Gemma-3) and inference budgets. Specifically, we consider two prompting strategies: direct assessment without intermediate reasoning tokens (``S'') and assessment with chain-of-thought tokens before the final prediction (``L''). \autoref{fig:generator_scaling} shows that our encoder matches the performance of the best-performing LLM judges in this setting, while remaining substantially less computationally expensive in terms of inference FLOPs.\footnote{Inference FLOPs are estimated using the formula from \citet{kaplan2020scaling}: FLOPs = 2 $\times$ model size (in parameters) $\times$ number of generated tokens.} Interestingly, we find that intermediate reasoning tokens do not improve LLM-based assessment accuracy, and that performance saturates around the 10B parameter scale, beyond which additional inference computation yields limited gains.

\section{Experimental Analysis}
\label{sec:xp_analysis}

In this section, we further investigate the properties of our encoder-based evaluation method through a series of complementary analyses.

\begin{wrapfigure}{r}{0.35\textwidth}
    \vspace{2pt}
    \centering
    \includegraphics[width=0.35\textwidth]{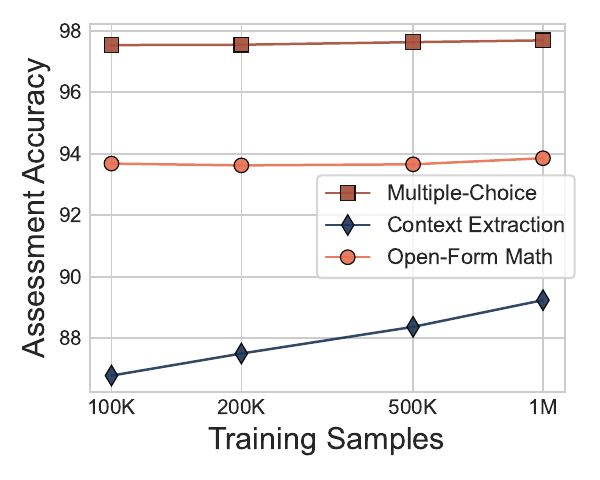}
    \vspace{-20pt}
    \caption{BERT-as-a-Judge evaluation quality across different training budgets.}
    \label{fig:sample_scaling}
    \vspace{-10pt}
\end{wrapfigure}

\paragraph{BERT-as-a-Judge is training-efficient.} By default, we train encoder models on 1M question-candidate-reference triplets. In this experiment, we evaluate lighter configurations: 500K, 200K, and 100K samples (\autoref{fig:sample_scaling}). Remarkably, 100K training samples are sufficient to accurately evaluate multiple-choice and open-form math tasks, with no significant improvement observed beyond this point. Gains are more noticeable for context extraction, as expected, 
since this task category requires more than simple candidate-reference matching and often demands understanding of the context provided by the question. Overall, with just 2 GPU hours of training (corresponding to the 100K-sample configuration), our encoder achieves a high assessment accuracy, making it well-suited for settings with limited data and computational resources.

\paragraph{Combining BERT-as-a-Judge with regex offers an efficient compromise.}
In this experiment (\autoref{tab:hybrid_approach}), we use BERT-as-a-Judge as a fallback when regex parsing fails. While it does not reach the performance of the standalone encoder, this hybrid approach substantially improves over regex alone. The results demonstrate that selectively applying the encoder can recover a significant portion of assessment accuracy while keeping computational overhead low (for example, reducing total compute by a factor of five for a model with 20\% regex failures).

\begin{table}[h]
    \centering
    \small
    \begin{minipage}{0.48\textwidth}
        \centering
        \small
        \resizebox{\textwidth}{!}{
        \begin{tabular}{l ccc}
\toprule
\multirow{2}{*}{\textbf{Task Category}} & \multirow{2}{*}{\textbf{Regex}} & \multirow{2}{*}{\textbf{BERTJ.}} & \textbf{Regex+} \\
& & & \textbf{BERTJ.} 
\\
\midrule
Multiple-Choice & 88.8 & \textbf{97.7} & 90.5 \\
Context Extraction & 73.0 & \textbf{89.2} & 75.2 \\
Open-Form Math & 87.3 & \textbf{93.9} & 89.9 \\
\bottomrule
\end{tabular}
        }
        \caption{Comparison of hybrid answer evaluation (Regex+BERT-J.) with standalone regex and BERT-as-a-Judge. Bold values indicate the highest accuracy in each row.}
        \label{tab:hybrid_approach}
    \end{minipage}
    \hfill
    \begin{minipage}{0.48\textwidth}
        \centering
        \small
        \resizebox{0.87\textwidth}{!}{
        \begin{tabular}{l ccc}
\toprule
\multirow{2}{*}{\textbf{Task Category}} & \multirow{2}{*}{\textbf{Regex}} & \multicolumn{2}{c}{\textbf{BERTJudge}} \\
\cmidrule(lr){3-4}
& & \textbf{w/ Q.} & \textbf{w/o Q.} \\
\midrule
Multiple-Choice & 88.8 & \textbf{97.7} & 97.3 \\
Context Extraction & 73.0 & \textbf{89.2} & 84.2 \\
Open-Form Math & 87.3 & \textbf{93.9} & \textbf{93.9} \\
\bottomrule
\end{tabular}
        }
        \caption{Impact of including (w/ Q.) or excluding (w/o Q.) the question in the encoder training prompt. Bold values denote the highest accuracy for each task category.}
        \label{tab:question_removal}
    \end{minipage}
\end{table}

\paragraph{Removing the question from the prompt yields a controlled performance decrease.}
As shown in \autoref{tab:question_removal}, omitting the question during encoder training (leaving only the candidate and reference) reduces overall assessment accuracy by removing some contextual information. However, this decrease is well-controlled across tasks. The question-free encoder still outperforms regex and remains close to the full-prompt setup, particularly on multiple-choice and open-form math tasks, while also reducing runtime due to shorter prompts and enabling application to any task with fully textual outputs, including multimodal tasks. The performance gap is slightly larger for context extraction, where the question provides critical information, underscoring the encoder’s context-aware capabilities.

\begin{table}[h]
\centering
\small
\resizebox{\textwidth}{!}{
\begin{NiceTabular}{l ccc || ccc}
\toprule
\multirow{2}{*}{\textbf{Task}} & \multicolumn{3}{c}{\textbf{Test Set {\scriptsize (Form.)}}} & \multicolumn{3}{c}{\textbf{Test Set {\scriptsize (Free)}}} \\
\cmidrule(lr){2-4} \cmidrule(lr){5-7}
 & \textbf{Regex} & \textbf{BERTJ. {\scriptsize (Form.)}} & \textbf{BERTJ. {\scriptsize (Free)}} & \textbf{Regex} & \textbf{BERTJ. {\scriptsize (Form.)}} & \textbf{BERTJ. {\scriptsize (Free)}} \\
\cmidrule(lr){1-4} \cmidrule(lr){5-8}
Multiple-Choice & 88.8 & 97.7 & 97.4 & -- & 94.0 & 97.6 \\
Context Extraction & 73.0 & 89.2 & 85.8 & -- & 84.3 & 91.6 \\
Open-Form Math & 87.3 & 93.9 & 93.7 & -- & 93.1 & 93.5 \\
\bottomrule
\end{NiceTabular}
}
\caption{Encoder's robustness to answer formatting. ``Test Set {\scriptsize (Form.)}'' denotes test sets with formatted answers, while ``Test Set {\scriptsize (Free)}'' contains unformatted answers. ``Regex'' columns show regex-based results, and ``BERT-J. {\scriptsize (Form.)}'' and ``BERT-J. {\scriptsize (Free)}'' report accuracies for BERT-as-a-Judge encoders trained on formatted and unformatted answers, respectively. Regex results are omitted for the free-format test set, as answers cannot be reliably parsed.}
\label{tab:format_robustness}
\end{table}

\paragraph{BERT-as-a-Judge is robust to variations in answer formatting guidelines.}
In our core experiments, and to ensure a fair comparison with the regex baseline, we train and evaluate the encoder on answers formatted to facilitate lexical parsing (\autoref{sec:dataset_generation}). In practice, however, users may follow custom guidelines or allow free-form responses.\footnote{Further details are provided in \autoref{sec:prompting_details}.} \autoref{tab:format_robustness} shows that under cross-formatting evaluation, meaning free-to-formatted (training on free-form, evaluating on formatted answers) and formatted-to-free (the reverse), we observe a slight performance drop compared to aligned settings. Nevertheless, the encoder still substantially outperforms regex, demonstrating strong robustness to variations in answer formatting. As expected, the free-to-formatted encoder consistently outperforms the formatted-to-free variant, benefiting from exposure to a wider range of formats during training and making it the preferred choice for downstream applications.

\begin{figure}[h]
    \centering
    \includegraphics[width=\textwidth]{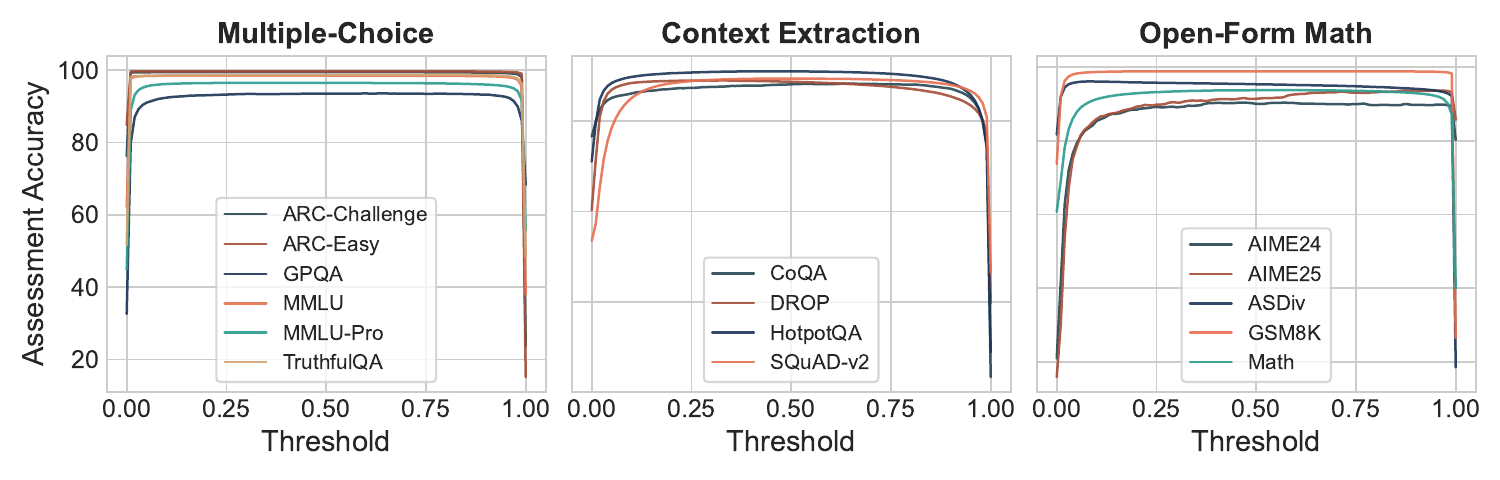}
    \vspace{-20pt}
    \caption{Effect of score thresholding on BERT-as-a-Judge downstream assessment accuracy across the three task categories, averaged over all models.}
    \label{fig:threshold_impact}
\end{figure}

\begin{wrapfigure}{r}{0.35\textwidth}
    \vspace{-8pt}
    \centering
    \includegraphics[width=0.35\textwidth]{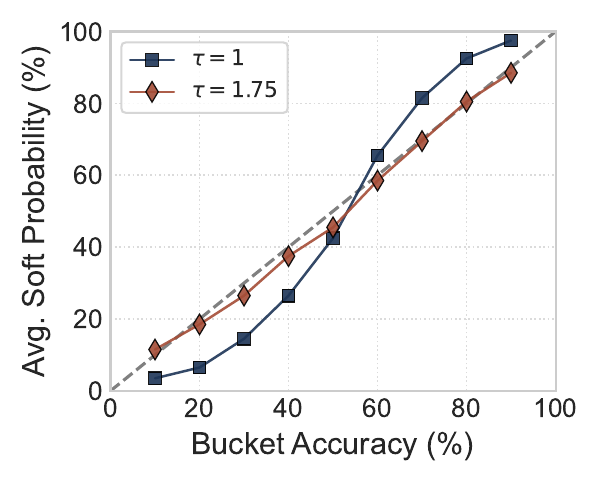}
    \vspace{-20pt}
    \caption{Calibration analysis of BERT-as-a-Judge's soft probabilities. The blue curve corresponds to the original (unscaled) soft probabilities, while the orange curve shows the probabilities after temperature scaling ($\tau$=1.75). The grey dashed line represents perfect calibration, where predicted probabilities match the true likelihood of the label being correct.}
    \label{fig:calibration}
    \vspace{-20pt}
\end{wrapfigure}

\paragraph{BERT-as-a-Judge is robust to decision threshold variations.}
Encoder classifiers output continuous sigmoid probabilities, requiring a decision threshold for discrete evaluation. While our main experiments (\autoref{sec:encoder_based_evaluation}) use a standard $0.5$ threshold, \autoref{fig:threshold_impact} demonstrates that accuracies remain remarkably stable across a broad spectrum of threshold values for all task categories. This invariance indicates strong separation between classes, enabling reliable off-the-shelf deployment without the need for task-specific threshold tuning.

\paragraph{BERT-as-a-Judge is well calibrated.}
Another desirable property of a classifier is the calibration of its soft probabilities, i.e., whether the predicted probabilities accurately reflect the true likelihood of the answer being correct. This is particularly important for confidence estimation. Empirically, calibration can be measured by grouping predictions into probability bins and comparing the average predicted probability with the empirical accuracy in each bin. A well-calibrated model should exhibit close agreement between these two quantities. \autoref{fig:calibration} shows that, while the raw soft probabilities are already reasonably calibrated, applying a modest temperature scaling ($\tau=1.75$) is sufficient to achieve near-perfect calibration.

\section{Related Work}

\paragraph{Traditional LLM evaluation.}
Pretrained language models have traditionally been evaluated using log-likelihood \citep{gpt2} or few-shot generation \citep{gpt3,gopher,palm,llama1,qwen1}. With the rise of instruction-tuned models, zero-shot generative evaluation has become standard \citep{wei2022finetuned,chung2024scaling,qwen3,eurollm22b}. This paradigm typically enforces structured outputs via prompting \citep{liang2023holistic,eval-harness}, followed by rule-based comparison to references using deterministic metrics such as exact match, ROUGE \citep{rouge}, Math-Verify \citep{mathverify}, or Code-Eval \citep{codeval}, making evaluation highly sensitive to surface-level formatting.

\paragraph{Model-based evaluation.}
Both lexical parsing and matching introduce limitations in capturing semantic correctness and robustness. Lexical overlap does not guarantee semantic equivalence, motivating neural metrics such as BERTScore \citep{bertscore} and InfoLM \citep{infolm} for general text generation, as well as task-specific evaluators like COMET \citep{comet,cometkiwi,xcomet}, MetricX \citep{metricx23,metricx25}, and BLEURT \citep{bleurt}. Additionally, reliably extracting model outputs is challenging when formatting is inconsistent \citep{ifeval,ifbench}. To mitigate these issues, LLM-as-a-Judge approaches \citep{zheng2023judging,wang2023judgelm,bavaresco2025llms,prometheus,prometheus2} directly assess candidate-reference equivalence across tasks, offering greater robustness to formatting artifacts, albeit at a substantial computational cost.

\section{Conclusion}

In this work, we show that standard evaluation protocols often conflate a model’s underlying problem-solving ability with its compliance to formatting constraints. Across extensive experiments spanning diverse models and tasks, we demonstrate that regex-based evaluation can substantially underestimate true performance. To address this, we propose BERT-as-a-Judge, a lightweight encoder-based framework that better captures semantic correctness, aligns more closely with human judgment, and avoids the high computational cost of LLM-as-a-Judge methods, enabling efficient and more reliable evaluation.

\section{Limitations and Future Work}

While BERT-as-a-Judge shows strong alignment with human judgments and effectively mitigates the limitations of lexical assessment, our study focuses on a specific subset of evaluation settings, namely, English benchmarks with objectively verifiable answers, where correctness can be clearly defined.

Building on these results and the existing literature, a natural next step is to broaden the scope of encoder-based evaluation toward more general-purpose settings. This includes expanding beyond fact-based and structured tasks to open-ended generation scenarios such as summarization, machine translation, code generation, and instruction following. Additionally, adapting the framework to multilingual contexts would further improve its applicability across diverse use cases. As foundation models continue to evolve toward multimodal capabilities, extending this approach to handle vision and speech inputs also presents a promising avenue. Exploring such cross-modal evaluation settings (e.g., visual question answering, image captioning, or speech-based reasoning) could help move toward a unified and efficient evaluation paradigm applicable across tasks and modalities.

\clearpage

\section*{Ethics Statement}

In conducting this research, we recognize the critical importance of fair and reliable evaluation in the LLM ecosystem. Evaluation metrics that are closely aligned with human judgments are essential to ensure that model comparisons accurately reflect real-world capabilities across the widest possible range of tasks. At the same time, the increasing scale of model evaluation, driven by more models, longer outputs, and a growing number of benchmark tasks, can impose substantial computational costs, raising concerns about accessibility and environmental impact. Our work emphasizes the development of lightweight, encoder-based evaluation methods that maintain high correlation with human judgments while minimizing compute requirements. By prioritizing both fairness and efficiency, we aim to support responsible, reproducible, and scalable evaluation practices in the broader LLM research community.

\section*{Acknowledgments}

We gratefully acknowledge the ADASTRA supercomputer at CINES for its technical support and access to HPC resources (grants C1615122 and GDA2401). This work was also supported by the French government under the France 2030 program (ArGiMi project). We further thank the 11 human annotators from the Artefact Research Center, whose careful efforts were instrumental in validating the correlation between our synthetic labeling strategy and human judgments, ensuring the reliability and rigor of our evaluation results.

\clearpage

\bibliography{colm2026_conference}
\bibliographystyle{colm2026_conference}

\clearpage

\appendix

\section{Prompting Details}
\label{sec:prompting_details}

In this section, we detail how model outputs are generated for both answer generation (\autoref{tab:generation_prompts}) and answer assessment (\autoref{tab:assessment_prompt_l} , \autoref{tab:assessment_prompt_s}). For answer generation, we also describe the suffixes used to impose different formatting instructions (\autoref{tab:generation_suffices}). Throughout the main text, we use the soft configuration by default, as it allows both regex-based answer parsing and the inclusion of intermediate chain-of-thought tokens, which improve answer quality (\autoref{sec:generation_mode}). For regex parsing under both the soft and strict formatting constraints, we use the pattern ``Final answer:\textbackslash\textbackslash s*(.+)'', which provides a general and flexible mechanism for extracting the predicted answer.

\begin{table}[h]
\centering
\small
\begin{tabular}{p{0.2\linewidth} p{0.55\linewidth}}
\toprule
\textbf{Task Category} & \textbf{Generation Prompt} \\
\midrule
\multirow{10}{*}{Multiple-Choice} & 
\begin{minipage}[t]{\linewidth}
\begin{verbatim}
Answer the following multiple-choice question.

Question: {question}

Choices:
A) {choice_1}
B) {choice_2}
C) {choice_3}
D) {choice_4}
[...]
\end{verbatim}
\end{minipage} \\
\midrule
\multirow{6}{*}{Context Extraction}  &
\begin{minipage}[t]{\linewidth}
\begin{verbatim}
Answer the question based on the provided context.

Context: 
{context}

Question: {question}
\end{verbatim}
\end{minipage} \\
\midrule
Open-Form Math &
\begin{minipage}[t]{\linewidth}
\begin{verbatim}
{question}
\end{verbatim}
\end{minipage} \\
\bottomrule
\end{tabular}
\caption{Base generation prompts for each task category.}
\label{tab:generation_prompts}
\end{table}

\begin{table}[h]
\centering
\small
\resizebox{\textwidth}{!}{
\begin{tabular}{p{0.2\linewidth}p{0.25\linewidth} p{0.5\linewidth}}
\toprule
\textbf{Task Category} & \textbf{Formatting Instruction} & \textbf{Generation Suffix} \\
\midrule

\multirow{6}{*}{Multiple-Choice} & Free &
\begin{minipage}[t]{\linewidth}
None
\end{minipage} \\
\addlinespace[3pt] \cdashline{2-3} \addlinespace[3pt]
 & \multirow{2}{*}{Soft} &
\begin{minipage}[t]{\linewidth}
\begin{verbatim}
Conclude your response with "Final answer: X", 
where X is the letter of the correct choice.
\end{verbatim}
\end{minipage} \\
\addlinespace[3pt] \cdashline{2-3} \addlinespace[3pt]
 & \multirow{3}{*}{Strict} &
\begin{minipage}[t]{\linewidth}
\begin{verbatim}
Respond only with the exact format "Final 
answer: X", where X is the letter of the correct 
choice.
\end{verbatim}
\end{minipage} \\
\midrule

\multirow{7}{*}{Context Extraction} & Free &
\begin{minipage}[t]{\linewidth}
None
\end{minipage} \\
\addlinespace[3pt] \cdashline{2-3} \addlinespace[3pt]
 & \multirow{3}{*}{Soft} &
\begin{minipage}[t]{\linewidth}
\begin{verbatim}
Conclude your response with "Final answer: X", 
where X is the exact span from the context that 
answers the question.
\end{verbatim}
\end{minipage} \\
\addlinespace[3pt] \cdashline{2-3} \addlinespace[3pt]
 & \multirow{3}{*}{Strict} &
\begin{minipage}[t]{\linewidth}
\begin{verbatim}
Respond only with the exact format "Final 
answer: X", where X is the exact span from the 
context that answers the question.
\end{verbatim}
\end{minipage} \\
\midrule

\multirow{5}{*}{Open-Form Math} & Free &
\begin{minipage}[t]{\linewidth}
None
\end{minipage} \\
\addlinespace[3pt] \cdashline{2-3} \addlinespace[3pt]
 & \multirow{2}{*}{Soft} &
\begin{minipage}[t]{\linewidth}
\begin{verbatim}
Conclude your response with "Final answer: X", 
where X is the computed solution.
\end{verbatim}
\end{minipage} \\
\addlinespace[3pt] \cdashline{2-3} \addlinespace[3pt]
 & \multirow{2}{*}{Strict} &
\begin{minipage}[t]{\linewidth}
\begin{verbatim}
Respond only with the exact format "Final 
answer: X", where X is the computed solution.
\end{verbatim}
\end{minipage} \\

\bottomrule
\end{tabular}
}
\caption{Generation suffixes used across task categories for all formatting strategies.}
\label{tab:generation_suffices}
\end{table}

\begin{table}[h]
\centering
\small
\resizebox{\textwidth}{!}{
\begin{tabular}{p{\linewidth}}
\toprule
\textbf{Assessment Prompt} \\
\midrule
\begin{minipage}[t]{\linewidth}
\begin{verbatim}
You are an expert evaluator. Your task is to determine whether the CANDIDATE response 
correctly answers the QUESTION.
Judge the CANDIDATE as correct only if its final answer, disregarding any intermediate 
reasoning or explanation, is semantically equivalent to the 
REFERENCE with respect to the QUESTION.
Base your judgment solely on the information given. Do not rely on external knowledge.

[QUESTION starts here]
{question}
[QUESTION ends here]

[REFERENCE starts here]
{reference}
[REFERENCE ends here]

[CANDIDATE starts here]
{candidate}
[CANDIDATE ends here]

Conclude your response with exactly one of the following:
- "Final answer: True" if the CANDIDATE is correct
- "Final answer: False" if the CANDIDATE is incorrect
\end{verbatim}
\end{minipage} \\
\bottomrule
\end{tabular}
}
\caption{Answer assessment prompt for LLM judges, allowing intermediate token generation before the final judgment. This involves Nemotron-Super-v1.5 for label generation (\autoref{sec:labeling}) and generative judges evaluated under inference budget L (\autoref{fig:generator_scaling}).}
\label{tab:assessment_prompt_l}
\end{table}

\begin{table}[h]
\centering
\small
\resizebox{\textwidth}{!}{
\begin{tabular}{p{\linewidth}}
\toprule
\textbf{Assessment Prompt} \\
\midrule
\begin{minipage}[t]{\linewidth}
\begin{verbatim}
You are an expert evaluator. Your task is to determine whether the CANDIDATE response 
correctly answers the QUESTION.
Judge the CANDIDATE as correct only if its final answer, disregarding any intermediate 
reasoning or explanation, is semantically equivalent to the 
REFERENCE with respect to the QUESTION.
Base your judgment solely on the information given. Do not rely on external knowledge.

[QUESTION starts here]
{question}
[QUESTION ends here]

[REFERENCE starts here]
{reference}
[REFERENCE ends here]

[CANDIDATE starts here]
{candidate}
[CANDIDATE ends here]

Respond with exactly one of the following strings (add no additional text):
- "Final answer: True" if the CANDIDATE is correct
- "Final answer: False" if the CANDIDATE is incorrect
\end{verbatim}
\end{minipage} \\
\bottomrule
\end{tabular}
}
\caption{Prompt used for direct assessment by LLM judges, applied to generative judges evaluated under inference budget S (\autoref{fig:generator_scaling}).}
\label{tab:assessment_prompt_s}
\end{table}


\clearpage

\section{Effect of Generation Mode on Downstream Performance}
\label{sec:generation_mode}

\begin{wraptable}{r}{0.45\textwidth}
\vspace{-12pt}
\centering
\small
\resizebox{0.45\textwidth}{!}{\begin{tabular}{lcccc}
\toprule
\multirow{2}{*}{\textbf{Task}} & \multirow{2}{*}{\textbf{Log-lik.}} & \multicolumn{3}{c}{\textbf{Generative}} \\
\cmidrule(lr){3-5}
 & & \textbf{Strict} & \textbf{Soft} & \textbf{Free} \\
\midrule
\multicolumn{5}{c}{\textit{Multiple-Choice}} \\
\addlinespace[2pt]
ARC-Challenge & 46.3 & 74.4 & \textbf{76.2} & 75.0 \\
ARC-Easy      & 65.0 & 84.6 & \textbf{84.9} & 84.4 \\
GPQA          & 25.9 & \textbf{32.3} & 31.8 & 28.8 \\
MMLU          & 39.9 & 59.4 & \textbf{62.0} & 60.5 \\
MMLU-Pro      & 21.1 & 38.0 & \textbf{44.3} & 42.6 \\
TruthfulQA    & 33.2 & 51.3 & \textbf{51.6} & 51.3 \\
\midrule
\multicolumn{5}{c}{\textit{Context Extraction}} \\
\addlinespace[2pt]
CoQA          & -- & 71.3 & 76.5 & \textbf{86.7} \\
DROP          & -- & 48.4 & 60.2 & \textbf{64.5} \\
HotpotQA      & -- & 66.0 & 71.0 & \textbf{82.4} \\
SQuAD-v2      & -- & 44.7 & 53.5 & \textbf{62.0} \\
\midrule
\multicolumn{5}{c}{\textit{Open-Form Math}} \\
\addlinespace[2pt]
AIME24        & -- & 18.1 & \textbf{19.8} & 18.1 \\
AIME25        & -- & 13.2 & 14.4 & \textbf{15.7} \\
ASDiv         & -- & 68.2 & 81.5 & \textbf{82.1} \\
GSM8K         & -- & 43.1 & 73.6 & \textbf{73.6} \\
Math          & -- & 49.3 & \textbf{60.2} & 57.3 \\
\bottomrule
\end{tabular}}
\caption{Comparison of evaluation modes across all benchmarks. Results are averaged over all models, and bold values indicate the best performance for each task.}
\label{tab:eval_mode_comparison}
\vspace{-35pt}
\end{wraptable}

In this section, we compare different answer production modes to assess their impact on model performance.  Specifically, answers are generated under three formatting regimes:\footnote{Full prompting details are provided in \autoref{sec:prompting_details}.} 
\begin{itemize}[leftmargin=15pt, itemsep=1pt, topsep=0pt]   
    \item \emph{Log-likelihood}: Candidate answers are iteratively appended to the prompt, and the model's prediction is derived from the sequence with the highest log-likelihood.\footnote{Applies only to multiple-choice tasks.}
    \item \emph{Strict}: The model is prompted to respond exactly with ``Final answer: [answer]''.
    \item \emph{Soft}: The model is prompted to conclude its response with ``Final answer: [answer]'' but may reason before answering.
    \item \emph{Free}: The model may answer in any format.
\end{itemize}

The results are reported in \autoref{tab:eval_mode_comparison}.

\paragraph{Models demonstrate greater capacity in generative mode.}
We first examine multiple-choice tasks by comparing generative evaluation against the log-likelihood approach. Our results indicate that the likelihood-based setup consistently and severely impairs performance across all evaluated multiple-choice benchmarks (e.g., -22.1\% on MMLU and -29.9\% on ARC-Challenge). This suggests that while likelihood evaluation offers a convenient, regex-free parsing mechanism, it significantly bottlenecks the model's inherent problem-solving capabilities compared to generative inference.

\paragraph{Strict formatting constraints impair performance.}
Setting aside likelihood-based evaluation, we compare strict and soft generative prompting strategies. We observe that strict prompting yields the lowest overall performance. While it performs comparably to the soft method on multiple-choice tasks, it significantly degrades performance on tasks requiring more complex outputs (e.g., -11.8\% on DROP and -30.5\% on GSM8K). This pronounced drop highlights the importance of allowing intermediate chain-of-thought generation to fully leverage the model’s problem-solving capacity.

\clearpage

\section{Human-Synthetic Label Agreement}
\label{sec:human_labels}

\begin{wraptable}[25]{r}{0.38\textwidth}
\vspace{-12pt}
\centering
\small
\begin{tabular}{lc}
\toprule
\textbf{Task} & \textbf{Accuracy (\%)} \\
\midrule
\multicolumn{2}{c}{\textit{Multiple-Choice}} \\
ARC-Challenge & 96.4 \\
ARC-Easy & 96.9 \\
MMLU & 96.4 \\
GPQA & 97.3 \\
MMLU-Pro & 97.5 \\
TruthfulQA & 96.3 \\
\midrule
\multicolumn{2}{c}{\textit{Context Extraction}} \\
HotpotQA & 96.7 \\
SQuAD-v2 & 97.1 \\
CoQA & 96.8 \\
DROP & 96.7 \\
\midrule
\multicolumn{2}{c}{\textit{Open-Form Math}} \\
GSM8K & 99.1 \\
Math & 98.6 \\
AIME24 & 98.7 \\
AIME25 & 99.4 \\
AsDiv & 97.7 \\
\bottomrule
\end{tabular}
\caption{Accuracy between human and synthetic labels across task categories.}
\label{tab:human_annotations}
\vspace{15pt}
\resizebox{0.38\textwidth}{!}{
\begin{tabular}{lc}
\toprule
\textbf{Error Category} & \textbf{Proportion (\%)} \\
\midrule
Interpretation & 33.0 \\
Verbose answer & 22.0 \\
Reading error (human) & 19.0 \\
Reading error (synthetic) & 17.0 \\
Semantic closeness & 9.0 \\
\midrule
\textbf{Total} & \textbf{100.0} \\
\bottomrule
\end{tabular}
}
\caption{Analysis of human-synthetic label disagreements.}
\label{tab:disagreement_analysis}
\end{wraptable}

\begin{figure}[h]
    \centering
    \includegraphics[width=\textwidth]{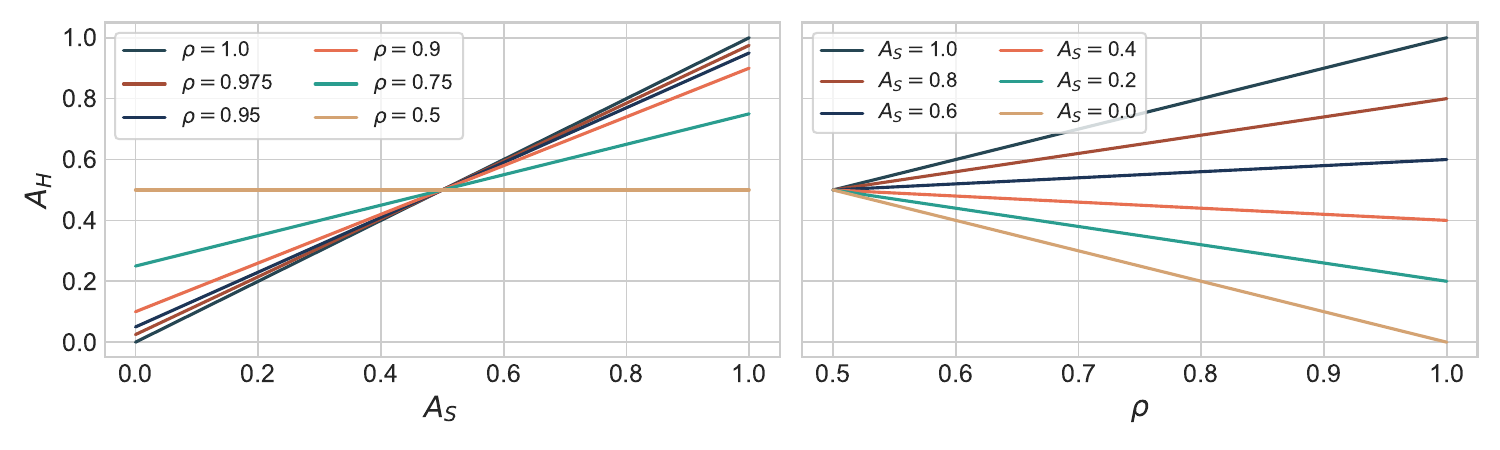}
    \vspace{-20pt}
    \caption{Sensitivity of the $A_H$ estimate to variations in $A_S$ and $\rho$.}
    \label{fig:accuracy_correction}
\end{figure}

This section complements \autoref{sec:labeling} of the main text by detailing the human annotation results. We first report the average agreement between human and synthetic evaluators across tasks (\autoref{tab:human_annotations}), followed by an in-depth analysis of disagreement cases (\autoref{tab:disagreement_analysis}). Finally, we evaluate how evaluator alignment impacts downstream performance measurements (\autoref{fig:accuracy_correction}).

\paragraph{Agreement measurement.} As shown in \autoref{tab:human_annotations}, we observe near-perfect consistency across all tasks: human and synthetic labels match on an average of 97.5\% of the 3,212 annotated samples. This strong alignment confirms the reliability of the synthetic judge, enabling us to scale assessments across large datasets without introducing systemic evaluation bias.

\paragraph{Discrepancy analysis.} To further validate this reliability, we examine the 80 disagreement cases (2.5\% of the annotated samples) detailed in \autoref{tab:disagreement_analysis}. We categorize these discrepancies into four primary types:
\begin{itemize}[leftmargin=15pt, itemsep=1pt, topsep=0pt]
    \item \emph{Interpretation:} Instances where the candidate's answer is inherently ambiguous and could reasonably be judged either way (e.g., due to over-answering or non-standard formatting).
    \item \emph{Verbose answers:} Lengthy responses containing multiple valid intermediate steps or sub-answers that confound the synthetic judge.
    \item \emph{Reading errors:} Straightforward misinterpretations committed by either the human or synthetic evaluator, with no obvious underlying textual artifact.
    \item \emph{Semantic closeness:} Discrepancies stemming from subtle semantic distinctions between reference and candidate answers, making fine-grained assessment challenging.
\end{itemize}

Crucially, severe discrepancies that could introduce systemic bias into the distillation process are rare. They account for less than half of all disagreements, representing under 1.25\% of the entire dataset. This confirms that training and evaluating with synthetic labels is robust and faithfully reflects true downstream performance.

\clearpage

\paragraph{Accuracy correction.} As described in \autoref{sec:assessment}, the reported accuracies in the main text are computed using synthetic labels generated by Nemotron-Super-v1.5. To estimate performance with respect to human annotations, we can apply a correction based on the observed agreement between human and synthetic labels. Let \(A_H\) denote the accuracy with respect to human labels (unknown), \(A_S\) the accuracy against synthetic labels, and \(\rho\) the agreement rate between synthetic and human judgments. Let \(Y_H\), \(Y_S\), and \(\hat{Y}\) be random variables representing, for a given example, the human label, the synthetic label, and the predicted label, respectively.

\begin{align}
A_H & = \mathbb{P} \left( \hat{Y} = Y_H \right) \\
    & = \mathbb{P} \left( \hat{Y} = Y_H | Y_H = Y_S \right) \mathbb{P} \left(Y_H = Y_S \right) + \mathbb{P} \left( \hat{Y} = Y_H | Y_H \neq Y_S \right) \mathbb{P} \left(Y_H \neq Y_S \right) \\
    & = \mathbb{P} \left( \hat{Y} = Y_S | Y_H = Y_S \right) \mathbb{P} \left(Y_H = Y_S \right) + \mathbb{P} \left( \hat{Y} \neq Y_S | Y_H \neq Y_S \right) \mathbb{P} \left(Y_H \neq Y_S \right) \\
    & = \mathbb{P} \left( \hat{Y} = Y_S \right) \mathbb{P} \left(Y_H = Y_S \right) + \ \mathbb{P} \left( \hat{Y} \neq Y_S \right) \mathbb{P} \left(Y_H \neq Y_S \right) \label{eq:independence} \\
    & = A_S\rho + (1 - A_S) (1- \rho) \\
    & = \left( 2 \rho - 1 \right) A_S + 1 - \rho \label{eq:final}
\end{align}

The final expression (\autoref{eq:final})\footnote{In \autoref{eq:independence}, we assume $ (\hat{Y}, Y_S) \perp (Y_H = Y_S) $, in line with our empirical observations. Intuitively, this means that agreement between the predicted and synthetic labels, $\hat{Y}$ and $Y_S$, is independent of whether the synthetic label $Y_S$ matches the human label $Y_H$.} can be interpreted as pulling the estimated accuracy $A_H$ toward random guessing. If human and synthetic labels are uncorrelated ($\rho = 0.5$), the estimated accuracy drops to a random guess, regardless of $A_S$ (see \autoref{fig:accuracy_correction} for illustration).

\clearpage

\section{Multilingual Results}
\label{sec:multilingual_results}

In this section, we complement the findings presented in the main text with preliminary results on the multilingual version of the MMLU benchmark. Specifically, we evaluate BERT-as-a-Judge under two settings: (1) using the English-only training setup described in the main text (\autoref{sec:evaluation_methods}), and (2) after a lightweight domain adaptation stage consisting of an additional fine-tuning phase on 20k multilingual question-candidate-reference triplets. We compare both configurations against Regex, Gemma-3 (12B), and Qwen-3 (14B). Results are presented in \autoref{tab:multilingual_results}.

\begin{table}[h]
\centering
\small
\begin{tabular}{lccccc}
    \toprule
    \multirow{2}{*}{\textbf{Language}} & \multirow{2}{*}{\textbf{Regex}} & \multirow{2}{*}{\textbf{Gemma-3 (12B)}} & \multirow{2}{*}{\textbf{Qwen-3 (14B)}} & \multicolumn{2}{c}{\textbf{BERTJudge}} \\
    \cmidrule(lr){5-6}
    & & & & \textbf{English} & \textbf{Multilingual} \\
    \midrule
    ar & 89.3 & 97.3 & 97.6 & 97.2 & \textbf{97.8} \\
    de & 87.9 & 96.9 & 96.8 & 96.5 & \textbf{97.5} \\
    en & 89.8 & 97.4 & 97.4 & 97.1 & \textbf{97.7} \\
    es & 89.7 & 97.2 & 97.2 & 96.3 & \textbf{97.7} \\
    fr & 88.4 & 96.8 & 96.8 & 96.2 & \textbf{97.6} \\
    hi & 88.1 & 95.9 & 95.4 & 97.2 & \textbf{97.5} \\
    it & 89.4 & 98.0 & \textbf{98.4} & 97.7 & \textbf{98.4} \\
    nl & 89.7 & 98.3 & \textbf{98.7} & 97.9 & 98.5 \\
    pt & 89.8 & 97.5 & 97.5 & 97.2 & \textbf{98.0} \\
    ru & 89.2 & 97.5 & 97.7 & 97.4 & \textbf{98.0} \\
    vi & 86.4 & 94.7 & 92.5 & 95.8 & \textbf{96.1} \\
    zh & 89.1 & 96.4 & 96.2 & 96.9 & \textbf{97.3} \\
    \bottomrule
\end{tabular}
\caption{Assessment accuracy across 12 languages of the multilingual MMLU benchmark. BERT-as-a-Judge results are reported for both the English-only fine-tuning setting and after multilingual adaptation. The highest accuracy in each row is shown in bold.}
\label{tab:multilingual_results}
\end{table}

BERT-as-a-Judge achieves strong assessment performance, matching or outperforming the much larger LLM judges evaluated. Interestingly, the initial English-only fine-tuning stage is already sufficient to obtain competitive performance, while a lightweight multilingual adaptation provides further incremental improvements in assessment accuracy.

\clearpage

\section{Detailed Results}
\label{sec:detailed_results}

\subsection{Regex Parsing Failures}

In this section, we extend \autoref{fig:regex_failures} from the main text by presenting disaggregated regex parsing failure rates across models and tasks.

\begin{table}[h]
\centering
\small
\begin{tabular}{llccccccc}
\toprule
\multirow{2}{*}{\textbf{Family}} & \multirow{2}{*}{\textbf{Size}} & \textbf{ARC} & \textbf{ARC} & \multirow{2}{*}{\textbf{GPQA}} & \multirow{2}{*}{\textbf{MMLU}} & \textbf{MMLU} & \textbf{Truthful} \\

 &  & \textbf{Challenge} & \textbf{Easy} &  &  & \textbf{Pro} & \textbf{QA} \\

\midrule

\multirow{2}{*}{Apertus}
 & 8B  & 0.2 & 0.1 & 4.9  & 0.8 & 4.5  & 0.0 \\
 & 70B & 0.1 & 0.1 & 21.2 & 2.6 & 12.3 & 1.5 \\

\midrule 

\multirow{3}{*}{EuroLLM}
 & 1.7B  & 97.6 & 98.7 & 83.0 & 94.1 & 69.9 & 66.3 \\
 & 9B    & 0.0  & 0.0  & 11.8 & 1.2  & 11.8 & 0.0 \\
 & 22B   & 0.1  & 0.0  & 7.4  & 0.4  & 3.0  & 0.0 \\

\midrule

\multirow{3}{*}{Falcon-3}
 & 1B & 5.1 & 4.0 & 11.2 & 5.5 & 31.9 & 3.7 \\
 & 3B & 0.0 & 0.2 & 2.0  & 0.2 & 2.5  & 0.1 \\
 & 7B & 0.0 & 0.0 & 1.1  & 0.1 & 1.3  & 0.1 \\

\midrule

\multirow{4}{*}{Gemma-3}
 & 1B  & 0.1 & 0.0 & 12.5 & 2.3 & 14.5 & 0.0 \\
 & 4B  & 0.0 & 0.0 & 6.2  & 0.3 & 3.3  & 0.0 \\
 & 12B & 0.0 & 0.0 & 4.9  & 0.1 & 3.0  & 0.0 \\
 & 27B & 0.0 & 0.0 & 3.6  & 0.1 & 1.3  & 0.0 \\

\midrule

\multirow{4}{*}{LFM-2}
 & 0.35B & 0.4 & 0.4 & 7.1  & 1.2 & 11.3 & 4.0 \\
 & 0.7B  & 2.9 & 2.8 & 18.1 & 6.1 & 15.5 & 1.6 \\
 & 1.2B  & 0.5 & 0.5 & 6.7  & 1.0 & 5.4  & 0.4 \\
 & 2.6B  & 0.0 & 0.2 & 10.0 & 1.4 & 9.9  & 0.6 \\

\midrule

\multirow{4}{*}{Llama-3}
 & 1B  & 5.0 & 3.6 & 36.4 & 9.0  & 26.0 & 5.4 \\
 & 3B  & 8.1 & 5.2 & 32.4 & 21.8 & 47.4 & 16.9 \\
 & 8B  & 0.1 & 0.3 & 25.2 & 2.6  & 16.1 & 2.3 \\
 & 70B & 0.0 & 0.0 & 7.4  & 0.9  & 6.0  & 0.0 \\

\midrule

\multirow{3}{*}{Ministral-3}
 & 3B  & 0.1 & 0.0 & 23.9 & 1.2  & 15.1 & 0.0 \\
 & 8B  & 0.0 & 0.0 & 25.0 & 1.4  & 10.3 & 0.0 \\
 & 14B & 0.2 & 0.1 & 16.5 & 1.3  & 8.6  & 0.0 \\

\midrule

\multirow{2}{*}{OLMo-3}
 & 7B  & 0.7 & 0.4 & 22.3 & 1.9  & 15.8 & 0.0 \\
 & 32B & 0.3 & 0.0 & 51.1 & 1.6  & 17.4 & 0.0 \\

\midrule

\multirow{2}{*}{Phi-4}
 & 3.6B  & 0.0 & 0.0 & 2.5  & 0.2  & 3.0 & 0.0 \\
 & 14B   & 0.0 & 0.0 & 0.9  & 0.0  & 0.6 & 0.0 \\

\midrule

\multirow{5}{*}{Qwen-3}
 & 0.6B  & 0.0 & 0.0 & 0.0  & 0.1  & 0.0 & 0.0 \\
 & 4B    & 0.0 & 0.0 & 9.6  & 0.4  & 4.6 & 0.0 \\
 & 8B    & 0.1 & 0.0 & 10.9 & 0.7  & 5.9 & 0.0 \\
 & 14B   & 0.1 & 0.0 & 11.4 & 0.3  & 4.7 & 0.0 \\
 & 32B   & 0.0 & 0.0 & 7.8  & 0.1  & 4.3 & 0.0 \\

\midrule

\multirow{4}{*}{SmolLM-2/3}
 & 0.135B & 98.4 & 98.4 & 97.8 & 97.6 & 97.8 & 98.3 \\
 & 0.36B  & 6.7  & 5.2  & 43.1 & 12.2 & 37.0 & 7.1 \\
 & 1.7B   & 0.0  & 0.0  & 0.2  & 0.2  & 0.5  & 0.2 \\
 & 3B     & 0.7  & 0.5  & 19.4 & 5.1  & 16.9 & 1.6 \\

\bottomrule
\end{tabular}
\caption{Parsing failure rates on multiple-choice benchmarks.}
\label{tab:detailed_results_parsing_failures_mc}
\end{table}

\begin{table}[h]
\centering
\small
\begin{tabular}{llcccc}
\toprule
\textbf{Family} & \textbf{Size} & \textbf{CoQA} & \textbf{DROP} & \textbf{HotpotQA} & \textbf{SQuAD-v2} \\
\midrule

\multirow{2}{*}{Apertus}
 & 8B   & 0.0 & 0.2 & 0.1 & 0.1 \\
 & 70B  & 0.0 & 0.3 & 0.0 & 0.1 \\

\midrule

\multirow{3}{*}{EuroLLM}
 & 1.7B  & 41.4 & 39.7 & 16.3 & 38.6 \\
 & 9B    & 0.0  & 0.0  & 0.0  & 0.0  \\
 & 22B   & 0.2  & 0.0  & 0.3  & 0.0  \\

\midrule

\multirow{3}{*}{Falcon-3}
 & 1B & 20.6 & 6.2 & 3.9 & 5.2 \\
 & 3B & 2.0  & 0.8 & 1.2 & 3.6 \\
 & 7B & 0.6  & 0.5 & 0.2 & 0.2 \\

\midrule

\multirow{4}{*}{Gemma-3}
 & 1B  & 0.0 & 0.8 & 0.1 & 0.3 \\
 & 4B  & 0.0 & 0.0 & 0.1 & 0.0 \\
 & 12B & 0.2 & 0.0 & 0.3 & 0.1 \\
 & 27B & 0.0 & 0.0 & 0.0 & 0.0 \\

\midrule

\multirow{4}{*}{LFM-2}
 & 0.35B & 1.8  & 2.2 & 4.1 & 3.5 \\
 & 0.7B  & 20.8 & 1.3 & 0.5 & 1.1 \\
 & 1.2B  & 0.2  & 0.5 & 1.5 & 0.2 \\
 & 2.6B  & 1.8  & 2.8 & 1.6 & 3.2 \\

\midrule

\multirow{4}{*}{Llama-3}
 & 1B  & 31.2 & 53.0 & 28.9 & 33.9 \\
 & 3B  & 34.6 & 5.3  & 17.1 & 27.9 \\
 & 8B  & 6.4  & 7.7  & 3.5  & 9.8  \\
 & 70B & 0.0  & 0.0  & 0.0  & 0.3  \\

\midrule

\multirow{3}{*}{Ministral-3}
 & 3B  & 0.0 & 0.1 & 0.1 & 0.1 \\
 & 8B  & 0.0 & 0.0 & 0.0 & 0.0 \\
 & 14B & 0.0 & 0.0 & 0.0 & 0.3 \\

\midrule

\multirow{2}{*}{OLMo-3}
 & 7B  & 0.0 & 2.2 & 0.5 & 1.8 \\
 & 32B & 0.0 & 0.1 & 0.1 & 2.3 \\

\midrule

\multirow{2}{*}{Phi-4}
 & 3.6B  & 0.0 & 0.0 & 0.0 & 0.0 \\
 & 14B   & 0.0 & 0.0 & 0.0 & 0.0 \\

\midrule

\multirow{5}{*}{Qwen-3}
 & 0.6B  & 0.0 & 0.0 & 0.0 & 0.0 \\
 & 4B    & 0.0 & 0.0 & 0.0 & 0.0 \\
 & 8B    & 0.0 & 0.1 & 0.0 & 0.0 \\
 & 14B   & 0.0 & 0.0 & 0.0 & 0.0 \\
 & 32B   & 0.0 & 0.0 & 0.0 & 0.0 \\

\midrule

\multirow{4}{*}{SmolLM-2/3}
 & 0.135B & 47.8 & 46.1 & 26.0 & 50.8 \\
 & 0.36B  & 54.2 & 44.2 & 24.8 & 48.1 \\
 & 1.7B   & 1.2  & 0.1  & 0.4  & 3.3  \\
 & 3B     & 0.4  & 0.7  & 0.1  & 0.3  \\

\bottomrule
\end{tabular}
\caption{Parsing failure rates on context extraction benchmarks.}
\label{tab:detailed_results_parsing_failures_ce}
\end{table}

\begin{table}[h]
\centering
\small
\begin{tabular}{llccccc}
\toprule
\textbf{Family} & \textbf{Size} & \textbf{AIME24} & \textbf{AIME25} & \textbf{ASDiv} & \textbf{GSM8K} & \textbf{Math} \\
\midrule

\multirow{2}{*}{Apertus}
 & 8B   & 50.0 & 46.7 & 0.9 & 0.8 & 13.3 \\
 & 70B  & 26.7 & 36.7 & 1.3 & 0.4 & 15.5 \\

\midrule

\multirow{3}{*}{EuroLLM}
 & 1.7B  & 70.0 & 76.7 & 5.3 & 6.3 & 54.1 \\
 & 9B    & 63.3 & 50.0 & 16.9 & 16.2 & 43.7 \\
 & 22B   & 33.3 & 30.0 & 2.5 & 6.8 & 11.7 \\

\midrule

\multirow{3}{*}{Falcon-3}
 & 1B & 83.3 & 90.0 & 69.0 & 59.9 & 77.7 \\
 & 3B & 46.7 & 40.0 & 0.9 & 0.5 & 16.2 \\
 & 7B & 13.3 & 16.7 & 0.0 & 0.2 & 2.9 \\

\midrule

\multirow{4}{*}{Gemma-3}
 & 1B  & 93.3 & 83.3 & 10.3 & 18.9 & 47.5 \\
 & 4B  & 50.0 & 16.7 & 0.3  & 0.7  & 8.9  \\
 & 12B & 33.3 & 33.3 & 0.1  & 0.2  & 9.4  \\
 & 27B & 33.3 & 23.3 & 0.2  & 0.2  & 5.2  \\

\midrule

\multirow{4}{*}{LFM-2}
 & 0.35B & 73.3 & 66.7 & 3.2 & 4.5 & 56.2 \\
 & 0.7B  & 60.0 & 66.7 & 1.3 & 2.8 & 29.3 \\
 & 1.2B  & 33.3 & 33.3 & 5.1 & 12.2 & 15.1 \\
 & 2.6B  & 33.3 & 43.3 & 0.3 & 0.4 & 13.3 \\

\midrule

\multirow{4}{*}{Llama-3}
 & 1B  & 100.0 & 100.0 & 25.9 & 28.1 & 84.9 \\
 & 3B  & 53.3  & 70.0  & 1.8  & 1.3  & 20.2 \\
 & 8B  & 53.3  & 56.7  & 3.1  & 1.2  & 18.5 \\
 & 70B & 100.0 & 100.0 & 6.7  & 25.4 & 92.1 \\

\midrule

\multirow{3}{*}{Ministral-3}
 & 3B  & 96.7 & 86.7 & 0.8 & 1.6 & 26.6 \\
 & 8B  & 86.7 & 90.0 & 0.7 & 1.4 & 19.1 \\
 & 14B & 93.3 & 86.7 & 0.4 & 1.1 & 23.4 \\

\midrule

\multirow{2}{*}{OLMo-3}
 & 7B  & 60.0 & 46.7 & 0.3 & 1.0 & 11.1 \\
 & 32B & 80.0 & 83.3 & 0.3 & 1.3 & 15.8 \\

\midrule

\multirow{2}{*}{Phi-4}
 & 3.6B  & 40.0 & 23.3 & 0.1 & 0.1 & 18.7 \\
 & 14B   & 16.7 & 16.7 & 0.0 & 0.0 & 3.2 \\

\midrule

\multirow{5}{*}{Qwen-3}
 & 0.6B  & 83.3 & 73.3 & 0.5 & 2.0 & 34.5 \\
 & 4B    & 56.7 & 63.3 & 0.2 & 0.6 & 8.7  \\
 & 8B    & 56.7 & 60.0 & 0.3 & 0.1 & 7.9  \\
 & 14B   & 50.0 & 46.7 & 0.1 & 0.0 & 6.1  \\
 & 32B   & 50.0 & 43.3 & 0.1 & 0.6 & 6.3  \\

\midrule

\multirow{4}{*}{SmolLM-2/3}
 & 0.135B & 70.0 & 66.7 & 36.4 & 40.6 & 68.2 \\
 & 0.36B  & 70.0 & 63.3 & 15.8 & 22.6 & 52.3 \\
 & 1.7B   & 63.3 & 80.0 & 2.4  & 5.8  & 38.7 \\
 & 3B     & 50.0 & 43.3 & 0.9  & 1.4  & 20.4 \\

\bottomrule
\end{tabular}
\caption{Parsing failure rates on open-form math benchmarks.}
\label{tab:detailed_results_parsing_failures_ofm}
\end{table}

\clearpage

\subsection{Impact of Regex-Based Evaluation on Performance Measurement}

This section extends \autoref{tab:regex_impact} from the main text by providing a breakdown of how regex-based evaluation affects downstream measured performance across every model and task.

\begin{table}[h]
\centering
\resizebox{\textwidth}{!}{
\begin{tabular}{llcccccccc}
\toprule
\multirow{2}{*}{\textbf{Family}} & \multirow{2}{*}{\textbf{Size}} & \multicolumn{2}{c}{\textbf{CoQA}} & \multicolumn{2}{c}{\textbf{DROP}} & \multicolumn{2}{c}{\textbf{HotpotQA}} & \multicolumn{2}{c}{\textbf{SQuAD-v2}} \\
\cmidrule(lr){3-4} \cmidrule(lr){5-6} \cmidrule(lr){7-8} \cmidrule(lr){9-10}
 & & \multicolumn{1}{c}{$\Delta$Accuracy} & \multicolumn{1}{c}{$\Delta$Rank} & \multicolumn{1}{c}{$\Delta$Accuracy} & \multicolumn{1}{c}{$\Delta$Rank} & \multicolumn{1}{c}{$\Delta$Accuracy} & \multicolumn{1}{c}{$\Delta$Rank} & \multicolumn{1}{c}{$\Delta$Accuracy} & \multicolumn{1}{c}{$\Delta$Rank} \\
\midrule
\multirow{2}{*}{Apertus} 
 & 8B & -30.2 {\tiny (-0.0/-30.2)} & \cellcolor{red!22}$\downarrow$3.0 & -21.8 {\tiny (-0.0/-21.8)} & \cellcolor{green!16}$\uparrow$1.5 & -20.6 {\tiny (-0.1/-20.5)} & \cellcolor{green!18}$\uparrow$2.0 & -19.2 {\tiny (-0.0/-19.2)} & \cellcolor{green!50}$\uparrow$10.0 \\
 & 70B & -41.4 {\tiny (-0.0/-41.4)} & \cellcolor{red!40}$\downarrow$7.5 & -21.9 {\tiny (-0.1/-21.9)} & \cellcolor{green!22}$\uparrow$3.0 & -22.7 {\tiny (-0.0/-22.7)} & \cellcolor{red!22}$\downarrow$3.0 & -25.3 {\tiny (-0.0/-25.3)} & \cellcolor{red!14}$\downarrow$1.0 \\
\midrule
\multirow{3}{*}{EuroLLM} 
 & 1.7B & -49.6 {\tiny (-30.6/-19.0)} & \cellcolor{red!42}$\downarrow$8.0 & -19.5 {\tiny (-11.6/-7.9)} & \cellcolor{red!14}$\downarrow$1.0 & -37.2 {\tiny (-9.9/-27.3)} & \cellcolor{red!34}$\downarrow$6.0 & -30.0 {\tiny (-16.6/-13.4)} & \cellcolor{red!38}$\downarrow$7.0 \\
 & 9B & -27.6 {\tiny (-0.0/-27.6)} & \cellcolor{red!14}$\downarrow$1.0 & -18.9 {\tiny (-0.0/-18.9)} & \cellcolor{green!14}$\uparrow$1.0 & -15.9 {\tiny (-0.0/-15.9)} & \cellcolor{green!30}$\uparrow$5.0 & -22.6 {\tiny (-0.0/-22.6)} & \cellcolor{green!34}$\uparrow$6.0 \\
 & 22B & -24.8 {\tiny (-0.2/-24.6)} & \cellcolor{green!14}$\uparrow$1.0 & -25.4 {\tiny (-0.0/-25.4)} & 0.0 & -16.3 {\tiny (-0.2/-16.1)} & \cellcolor{green!14}$\uparrow$1.0 & -29.2 {\tiny (-0.0/-29.2)} & 0.0 \\
\midrule
\multirow{3}{*}{Falcon-3} 
 & 1B & -42.6 {\tiny (-16.2/-26.4)} & \cellcolor{red!40}$\downarrow$7.5 & -11.5 {\tiny (-1.7/-9.8)} & 0.0 & -17.4 {\tiny (-2.2/-15.3)} & 0.0 & -19.3 {\tiny (-2.1/-17.2)} & 0.0 \\
 & 3B & -34.0 {\tiny (-2.0/-32.0)} & \cellcolor{red!14}$\downarrow$1.0 & -18.6 {\tiny (-0.5/-18.1)} & 0.0 & -23.4 {\tiny (-1.0/-22.4)} & \cellcolor{red!14}$\downarrow$1.0 & -21.7 {\tiny (-1.7/-20.0)} & \cellcolor{green!18}$\uparrow$2.0 \\
 & 7B & -19.8 {\tiny (-0.4/-19.4)} & \cellcolor{green!26}$\uparrow$4.0 & -14.6 {\tiny (-0.3/-14.3)} & \cellcolor{green!34}$\uparrow$6.0 & -12.4 {\tiny (-0.2/-12.2)} & \cellcolor{green!26}$\uparrow$4.0 & -18.0 {\tiny (-0.1/-17.9)} & \cellcolor{green!66}$\uparrow$14.0 \\
\midrule
\multirow{4}{*}{Gemma-3} 
 & 1B & -19.0 {\tiny (-0.0/-19.0)} & \cellcolor{green!52}$\uparrow$10.5 & -6.0 {\tiny (-0.1/-5.9)} & \cellcolor{green!18}$\uparrow$2.0 & -9.9 {\tiny (-0.1/-9.8)} & \cellcolor{green!18}$\uparrow$2.0 & -14.1 {\tiny (-0.1/-14.0)} & \cellcolor{green!26}$\uparrow$4.0 \\
 & 4B & -43.0 {\tiny (-0.0/-43.0)} & \cellcolor{red!30}$\downarrow$5.0 & -27.4 {\tiny (-0.0/-27.4)} & \cellcolor{green!18}$\uparrow$2.0 & -23.8 {\tiny (-0.1/-23.8)} & 0.0 & -29.0 {\tiny (-0.0/-28.9)} & \cellcolor{red!22}$\downarrow$3.0 \\
 & 12B & -28.8 {\tiny (-0.2/-28.6)} & \cellcolor{red!30}$\downarrow$5.0 & -31.8 {\tiny (-0.0/-31.8)} & \cellcolor{red!30}$\downarrow$5.0 & -20.1 {\tiny (-0.3/-19.9)} & \cellcolor{red!38}$\downarrow$7.0 & -29.5 {\tiny (-0.1/-29.4)} & 0.0 \\
 & 27B & -28.8 {\tiny (-0.0/-28.8)} & \cellcolor{red!14}$\downarrow$1.0 & -31.1 {\tiny (-0.0/-31.1)} & \cellcolor{red!26}$\downarrow$4.0 & -25.4 {\tiny (-0.0/-25.4)} & \cellcolor{red!20}$\downarrow$2.5 & -29.4 {\tiny (-0.0/-29.4)} & \cellcolor{red!14}$\downarrow$1.0 \\
\midrule
\multirow{4}{*}{LFM-2} 
 & 0.35B & -20.0 {\tiny (-1.4/-18.6)} & \cellcolor{green!44}$\uparrow$8.5 & -3.7 {\tiny (-0.6/-3.1)} & \cellcolor{green!22}$\uparrow$3.0 & -8.2 {\tiny (-2.0/-6.1)} & \cellcolor{green!22}$\uparrow$3.0 & -10.3 {\tiny (-1.5/-8.8)} & \cellcolor{green!30}$\uparrow$5.0 \\
 & 0.7B & -33.6 {\tiny (-15.8/-17.8)} & \cellcolor{green!14}$\uparrow$1.0 & -5.5 {\tiny (-0.4/-5.1)} & \cellcolor{green!18}$\uparrow$2.0 & -10.3 {\tiny (-0.3/-10.0)} & \cellcolor{green!18}$\uparrow$2.0 & -11.4 {\tiny (-0.6/-10.7)} & \cellcolor{green!22}$\uparrow$3.0 \\
 & 1.2B & -14.8 {\tiny (-0.0/-14.8)} & \cellcolor{green!68}$\uparrow$14.5 & -8.9 {\tiny (-0.1/-8.8)} & \cellcolor{green!18}$\uparrow$2.0 & -9.9 {\tiny (-1.2/-8.7)} & \cellcolor{green!38}$\uparrow$7.0 & -13.2 {\tiny (-0.0/-13.1)} & \cellcolor{green!62}$\uparrow$13.0 \\
 & 2.6B & -29.6 {\tiny (-1.6/-28.0)} & \cellcolor{green!20}$\uparrow$2.5 & -28.6 {\tiny (-1.2/-27.4)} & \cellcolor{red!18}$\downarrow$2.0 & -22.6 {\tiny (-1.1/-21.5)} & 0.0 & -33.5 {\tiny (-1.2/-32.4)} & \cellcolor{red!46}$\downarrow$9.0 \\
\midrule
\multirow{4}{*}{Llama-3} 
 & 1B & -30.0 {\tiny (-21.0/-9.0)} & \cellcolor{green!22}$\uparrow$3.0 & -24.1 {\tiny (-20.4/-3.7)} & \cellcolor{red!34}$\downarrow$6.0 & -22.0 {\tiny (-17.2/-4.8)} & 0.0 & -19.7 {\tiny (-13.9/-5.8)} & \cellcolor{red!18}$\downarrow$2.0 \\
 & 3B & -40.4 {\tiny (-29.2/-11.2)} & \cellcolor{red!28}$\downarrow$4.5 & -17.9 {\tiny (-3.0/-14.9)} & \cellcolor{green!42}$\uparrow$8.0 & -22.6 {\tiny (-13.4/-9.3)} & 0.0 & -29.1 {\tiny (-17.0/-12.1)} & \cellcolor{red!30}$\downarrow$5.0 \\
 & 8B & -23.6 {\tiny (-4.6/-19.0)} & \cellcolor{red!14}$\downarrow$1.0 & -21.5 {\tiny (-3.7/-17.8)} & \cellcolor{green!22}$\uparrow$3.0 & -17.1 {\tiny (-2.0/-15.1)} & \cellcolor{green!22}$\uparrow$3.0 & -21.9 {\tiny (-5.9/-16.0)} & \cellcolor{green!46}$\uparrow$9.0 \\
 & 70B & -17.2 {\tiny (-0.0/-17.2)} & \cellcolor{green!38}$\uparrow$7.0 & -17.0 {\tiny (-0.0/-17.0)} & \cellcolor{green!22}$\uparrow$3.0 & -11.5 {\tiny (-0.0/-11.5)} & \cellcolor{green!50}$\uparrow$10.0 & -27.8 {\tiny (-0.3/-27.5)} & \cellcolor{green!34}$\uparrow$6.0 \\
\midrule
\multirow{3}{*}{Ministral} 
 & 3B & -36.8 {\tiny (-0.0/-36.8)} & \cellcolor{red!14}$\downarrow$1.0 & -32.9 {\tiny (-0.0/-32.9)} & \cellcolor{red!34}$\downarrow$6.0 & -29.8 {\tiny (-0.0/-29.7)} & \cellcolor{red!30}$\downarrow$5.0 & -30.6 {\tiny (-0.0/-30.6)} & \cellcolor{red!42}$\downarrow$8.0 \\
 & 8B & -30.0 {\tiny (-0.0/-30.0)} & \cellcolor{green!18}$\uparrow$2.0 & -35.0 {\tiny (-0.0/-35.0)} & \cellcolor{red!46}$\downarrow$9.0 & -24.9 {\tiny (-0.0/-24.9)} & \cellcolor{red!34}$\downarrow$6.0 & -34.0 {\tiny (-0.0/-34.0)} & \cellcolor{red!34}$\downarrow$6.0 \\
 & 14B & -27.4 {\tiny (-0.0/-27.4)} & \cellcolor{red!18}$\downarrow$2.0 & -30.7 {\tiny (-0.0/-30.7)} & \cellcolor{red!30}$\downarrow$5.0 & -23.4 {\tiny (-0.0/-23.4)} & \cellcolor{red!46}$\downarrow$9.0 & -36.3 {\tiny (-0.0/-36.3)} & \cellcolor{red!38}$\downarrow$7.0 \\
\midrule
\multirow{2}{*}{OLMo-3} 
 & 7B & -17.6 {\tiny (-0.0/-17.6)} & \cellcolor{green!30}$\uparrow$5.0 & -22.5 {\tiny (-0.8/-21.7)} & \cellcolor{green!26}$\uparrow$4.0 & -11.7 {\tiny (-0.0/-11.6)} & \cellcolor{green!48}$\uparrow$9.5 & -26.4 {\tiny (-0.2/-26.2)} & \cellcolor{green!38}$\uparrow$7.0 \\
 & 32B & -24.8 {\tiny (-0.0/-24.8)} & \cellcolor{red!24}$\downarrow$3.5 & -27.3 {\tiny (-0.0/-27.2)} & \cellcolor{red!18}$\downarrow$2.0 & -16.1 {\tiny (-0.0/-16.0)} & 0.0 & -35.6 {\tiny (-2.2/-33.4)} & \cellcolor{red!26}$\downarrow$4.0 \\
\midrule
\multirow{2}{*}{Phi-4} 
 & 3.6B & -42.0 {\tiny (-0.0/-42.0)} & \cellcolor{red!44}$\downarrow$8.5 & -23.2 {\tiny (-0.0/-23.2)} & \cellcolor{green!14}$\uparrow$1.0 & -29.4 {\tiny (-0.0/-29.4)} & \cellcolor{red!22}$\downarrow$3.0 & -35.1 {\tiny (-0.0/-35.1)} & \cellcolor{red!58}$\downarrow$12.0 \\
 & 14B & -36.0 {\tiny (-0.0/-36.0)} & \cellcolor{red!12}$\downarrow$0.5 & -32.0 {\tiny (-0.0/-32.0)} & \cellcolor{red!26}$\downarrow$4.0 & -27.8 {\tiny (-0.0/-27.8)} & \cellcolor{red!42}$\downarrow$8.0 & -45.5 {\tiny (-0.0/-45.5)} & \cellcolor{red!82}$\downarrow$18.0 \\
\midrule
\multirow{5}{*}{Qwen-3} 
 & 0.6B & -27.2 {\tiny (-0.0/-27.2)} & \cellcolor{green!26}$\uparrow$4.0 & -9.4 {\tiny (-0.0/-9.4)} & \cellcolor{green!14}$\uparrow$1.0 & -7.3 {\tiny (-0.0/-7.3)} & \cellcolor{green!22}$\uparrow$3.0 & -11.1 {\tiny (-0.0/-11.1)} & \cellcolor{green!22}$\uparrow$3.0 \\
 & 4B & -23.0 {\tiny (-0.0/-23.0)} & \cellcolor{green!34}$\uparrow$6.0 & -20.0 {\tiny (-0.0/-20.0)} & \cellcolor{green!34}$\uparrow$6.0 & -12.9 {\tiny (-0.0/-12.9)} & \cellcolor{green!38}$\uparrow$7.0 & -25.0 {\tiny (-0.0/-25.0)} & \cellcolor{green!46}$\uparrow$9.0 \\
 & 8B & -30.6 {\tiny (-0.0/-30.6)} & \cellcolor{red!42}$\downarrow$8.0 & -34.5 {\tiny (-0.0/-34.4)} & \cellcolor{red!46}$\downarrow$9.0 & -19.7 {\tiny (-0.0/-19.7)} & \cellcolor{red!26}$\downarrow$4.0 & -34.2 {\tiny (-0.0/-34.2)} & \cellcolor{red!42}$\downarrow$8.0 \\
 & 14B & -19.8 {\tiny (-0.0/-19.8)} & 0.0 & -19.9 {\tiny (-0.0/-19.9)} & \cellcolor{red!14}$\downarrow$1.0 & -12.8 {\tiny (-0.0/-12.8)} & 0.0 & -31.1 {\tiny (-0.0/-31.1)} & \cellcolor{green!22}$\uparrow$3.0 \\
 & 32B & -23.0 {\tiny (-0.0/-23.0)} & \cellcolor{red!20}$\downarrow$2.5 & -25.3 {\tiny (-0.0/-25.3)} & \cellcolor{green!18}$\uparrow$2.0 & -17.1 {\tiny (-0.0/-17.1)} & \cellcolor{red!18}$\downarrow$2.0 & -35.1 {\tiny (-0.0/-35.1)} & \cellcolor{red!18}$\downarrow$2.0 \\
\midrule
\multirow{4}{*}{SmolLM-2/3} 
 & 0.135B & -38.0 {\tiny (-24.2/-13.8)} & 0.0 & -11.5 {\tiny (-6.8/-4.7)} & \cellcolor{green!14}$\uparrow$1.0 & -31.7 {\tiny (-10.1/-21.6)} & \cellcolor{red!14}$\downarrow$1.0 & -23.9 {\tiny (-12.5/-11.4)} & 0.0 \\
 & 0.36B & -55.0 {\tiny (-35.2/-19.8)} & \cellcolor{red!34}$\downarrow$6.0 & -17.6 {\tiny (-11.4/-6.1)} & \cellcolor{red!18}$\downarrow$2.0 & -36.5 {\tiny (-12.7/-23.8)} & \cellcolor{red!26}$\downarrow$4.0 & -27.9 {\tiny (-16.0/-11.9)} & \cellcolor{red!22}$\downarrow$3.0 \\
 & 1.7B & -25.2 {\tiny (-1.0/-24.2)} & \cellcolor{green!36}$\uparrow$6.5 & -8.4 {\tiny (-0.1/-8.3)} & 0.0 & -14.7 {\tiny (-0.3/-14.4)} & \cellcolor{green!22}$\uparrow$3.0 & -18.3 {\tiny (-1.5/-16.8)} & 0.0 \\
 & 3B & -35.2 {\tiny (-0.4/-34.8)} & \cellcolor{green!14}$\uparrow$1.0 & -20.5 {\tiny (-0.1/-20.3)} & \cellcolor{green!28}$\uparrow$4.5 & -22.3 {\tiny (-0.0/-22.2)} & 0.0 & -25.4 {\tiny (-0.0/-25.3)} & \cellcolor{green!18}$\uparrow$2.0 \\
\bottomrule
\end{tabular}
}
\caption{Effect of regex-based evaluation on performance measurement for context extraction benchmarks.}
\label{tab:detailed_results_regex_delta_ce}
\end{table}

\begin{table}[h]
\centering
\small
\resizebox{\textwidth}{!}{
\begin{tabular}{llcccccccccccc}
\toprule
\multirow{2}{*}{\textbf{Family}} & \multirow{2}{*}{\textbf{Size}} & \multicolumn{2}{c}{\textbf{ARC-Challenge}} & \multicolumn{2}{c}{\textbf{ARC-Easy}} & \multicolumn{2}{c}{\textbf{GPQA}} & \multicolumn{2}{c}{\textbf{MMLU}} & \multicolumn{2}{c}{\textbf{MMLU-Pro}} & \multicolumn{2}{c}{\textbf{TruthfulQA}} \\
\cmidrule(lr){3-4} \cmidrule(lr){5-6} \cmidrule(lr){7-8} \cmidrule(lr){9-10} \cmidrule(lr){11-12} \cmidrule(lr){13-14}
 & & \multicolumn{1}{c}{$\Delta$Accuracy} & \multicolumn{1}{c}{$\Delta$Rank} & \multicolumn{1}{c}{$\Delta$Accuracy} & \multicolumn{1}{c}{$\Delta$Rank} & \multicolumn{1}{c}{$\Delta$Accuracy} & \multicolumn{1}{c}{$\Delta$Rank} & \multicolumn{1}{c}{$\Delta$Accuracy} & \multicolumn{1}{c}{$\Delta$Rank} & \multicolumn{1}{c}{$\Delta$Accuracy} & \multicolumn{1}{c}{$\Delta$Rank} & \multicolumn{1}{c}{$\Delta$Accuracy} & \multicolumn{1}{c}{$\Delta$Rank} \\
\midrule
\multirow{2}{*}{Apertus} 
 & 8B & 0.0 {\tiny (-0.1/+0.1)} & \cellcolor{green!31}$\uparrow$3.0 & 0.0 {\tiny (-0.0/0.0)} & \cellcolor{green!52}$\uparrow$6.0 & +5.8 {\tiny (-0.0/+5.8)} & \cellcolor{green!100}$\uparrow$16.0 & -0.9 {\tiny (-0.1/-0.8)} & \cellcolor{green!38}$\uparrow$4.0 & +0.9 {\tiny (-0.5/+1.4)} & \cellcolor{green!38}$\uparrow$4.0 & -0.5 {\tiny (-0.0/-0.5)} & 0.0 \\
 & 70B & -1.8 {\tiny (-0.1/-1.7)} & \cellcolor{green!31}$\uparrow$3.0 & -1.5 {\tiny (-0.0/-1.5)} & \cellcolor{green!24}$\uparrow$2.0 & -1.3 {\tiny (-1.3/0.0)} & \cellcolor{green!70}$\uparrow$8.5 & -3.5 {\tiny (-0.4/-3.1)} & \cellcolor{green!31}$\uparrow$3.0 & -2.9 {\tiny (-1.2/-1.7)} & \cellcolor{green!17}$\uparrow$1.0 & -3.3 {\tiny (-0.5/-2.8)} & \cellcolor{red!16}$\downarrow$2.0 \\
\midrule
\multirow{3}{*}{EuroLLM} 
 & 1.7B & -29.8 {\tiny (-28.2/-1.5)} & \cellcolor{red!13}$\downarrow$1.0 & -29.2 {\tiny (-28.4/-0.8)} & \cellcolor{red!13}$\downarrow$1.0 & -28.3 {\tiny (-19.2/-9.2)} & \cellcolor{red!63}$\downarrow$17.5 & -27.4 {\tiny (-24.6/-2.9)} & \cellcolor{red!13}$\downarrow$1.0 & -22.4 {\tiny (-9.3/-13.1)} & \cellcolor{red!31}$\downarrow$7.0 & -29.0 {\tiny (-16.3/-12.7)} & \cellcolor{red!25}$\downarrow$5.0 \\
 & 9B & -0.3 {\tiny (-0.0/-0.3)} & \cellcolor{green!38}$\uparrow$4.0 & -0.0 {\tiny (-0.0/-0.0)} & \cellcolor{green!52}$\uparrow$6.0 & -8.9 {\tiny (-2.7/-6.3)} & \cellcolor{green!14}$\uparrow$0.5 & -1.6 {\tiny (-0.5/-1.2)} & \cellcolor{green!24}$\uparrow$2.0 & -8.2 {\tiny (-2.7/-5.5)} & 0.0 & +0.2 {\tiny (-0.0/+0.2)} & \cellcolor{green!31}$\uparrow$3.0 \\
 & 22B & -8.6 {\tiny (-0.1/-8.5)} & \cellcolor{red!28}$\downarrow$6.0 & -7.7 {\tiny (-0.0/-7.7)} & \cellcolor{red!34}$\downarrow$8.0 & -9.4 {\tiny (-1.1/-8.3)} & \cellcolor{red!18}$\downarrow$2.5 & -12.7 {\tiny (-0.1/-12.6)} & \cellcolor{red!34}$\downarrow$8.0 & -10.0 {\tiny (-0.7/-9.3)} & \cellcolor{red!16}$\downarrow$2.0 & -6.2 {\tiny (-0.0/-6.2)} & \cellcolor{red!13}$\downarrow$1.0 \\
\midrule
\multirow{3}{*}{Falcon-3} 
 & 1B & -26.7 {\tiny (-2.6/-24.1)} & 0.0 & -33.1 {\tiny (-3.0/-30.1)} & 0.0 & -14.7 {\tiny (-2.2/-12.5)} & \cellcolor{red!42}$\downarrow$10.5 & -15.4 {\tiny (-1.7/-13.7)} & \cellcolor{green!24}$\uparrow$2.0 & -13.8 {\tiny (-6.3/-7.5)} & 0.0 & -6.6 {\tiny (-0.6/-6.0)} & \cellcolor{green!31}$\uparrow$3.0 \\
 & 3B & -3.2 {\tiny (-0.0/-3.2)} & \cellcolor{green!24}$\uparrow$2.0 & -2.2 {\tiny (-0.1/-2.1)} & \cellcolor{green!24}$\uparrow$2.0 & -4.5 {\tiny (-0.2/-4.2)} & \cellcolor{green!59}$\uparrow$7.0 & -3.9 {\tiny (-0.1/-3.9)} & \cellcolor{green!24}$\uparrow$2.0 & -7.5 {\tiny (-0.4/-7.1)} & \cellcolor{red!13}$\downarrow$1.0 & -1.1 {\tiny (-0.1/-1.0)} & \cellcolor{green!24}$\uparrow$2.0 \\
 & 7B & -5.2 {\tiny (-0.0/-5.2)} & \cellcolor{red!16}$\downarrow$2.0 & -6.6 {\tiny (-0.0/-6.6)} & \cellcolor{red!22}$\downarrow$4.0 & -6.0 {\tiny (-0.7/-5.4)} & \cellcolor{green!24}$\uparrow$2.0 & -6.7 {\tiny (-0.0/-6.7)} & \cellcolor{red!13}$\downarrow$1.0 & -7.1 {\tiny (-0.4/-6.7)} & \cellcolor{green!17}$\uparrow$1.0 & -2.0 {\tiny (-0.0/-2.0)} & \cellcolor{green!17}$\uparrow$1.0 \\
\midrule
\multirow{4}{*}{Gemma-3} 
 & 1B & +0.1 {\tiny (-0.1/+0.2)} & \cellcolor{green!45}$\uparrow$5.0 & -0.6 {\tiny (-0.0/-0.6)} & \cellcolor{green!45}$\uparrow$5.0 & +2.5 {\tiny (-1.6/+4.0)} & \cellcolor{green!87}$\uparrow$11.0 & -1.0 {\tiny (-0.6/-0.4)} & \cellcolor{green!38}$\uparrow$4.0 & -2.2 {\tiny (-1.8/-0.5)} & \cellcolor{green!45}$\uparrow$5.0 & +0.5 {\tiny (-0.0/+0.5)} & \cellcolor{green!38}$\uparrow$4.0 \\
 & 4B & +0.2 {\tiny (-0.0/+0.2)} & \cellcolor{green!31}$\uparrow$3.0 & 0.0 {\tiny (-0.0/0.0)} & \cellcolor{green!52}$\uparrow$6.0 & +1.1 {\tiny (-0.4/+1.6)} & \cellcolor{green!94}$\uparrow$12.0 & +0.9 {\tiny (-0.1/+0.9)} & \cellcolor{green!59}$\uparrow$7.0 & -0.4 {\tiny (-0.5/+0.1)} & \cellcolor{green!52}$\uparrow$6.0 & +0.2 {\tiny (-0.0/+0.2)} & \cellcolor{green!31}$\uparrow$3.0 \\
 & 12B & +0.3 {\tiny (-0.0/+0.3)} & \cellcolor{green!52}$\uparrow$6.0 & 0.0 {\tiny (-0.0/0.0)} & \cellcolor{green!38}$\uparrow$4.0 & -0.4 {\tiny (-0.4/0.0)} & \cellcolor{green!59}$\uparrow$7.0 & 0.0 {\tiny (-0.0/0.0)} & \cellcolor{green!52}$\uparrow$6.0 & -1.1 {\tiny (-0.5/-0.6)} & \cellcolor{green!45}$\uparrow$5.0 & +0.6 {\tiny (-0.0/+0.6)} & \cellcolor{green!38}$\uparrow$4.0 \\
 & 27B & +0.3 {\tiny (-0.0/+0.3)} & \cellcolor{green!45}$\uparrow$5.0 & +0.2 {\tiny (-0.0/+0.2)} & \cellcolor{green!38}$\uparrow$4.0 & +0.2 {\tiny (-0.2/+0.4)} & \cellcolor{green!52}$\uparrow$6.0 & +0.3 {\tiny (-0.0/+0.3)} & \cellcolor{green!38}$\uparrow$4.0 & -0.9 {\tiny (-0.4/-0.5)} & \cellcolor{green!45}$\uparrow$5.0 & +0.2 {\tiny (-0.0/+0.2)} & \cellcolor{green!38}$\uparrow$4.0 \\
\midrule
\multirow{4}{*}{LFM-2} 
 & 0.35B & -36.1 {\tiny (-0.2/-35.9)} & \cellcolor{green!17}$\uparrow$1.0 & -50.0 {\tiny (-0.4/-49.6)} & \cellcolor{green!17}$\uparrow$1.0 & -9.8 {\tiny (-0.4/-9.4)} & \cellcolor{green!35}$\uparrow$3.5 & -29.9 {\tiny (-0.5/-29.4)} & \cellcolor{green!17}$\uparrow$1.0 & -15.0 {\tiny (-1.8/-13.2)} & 0.0 & -20.3 {\tiny (-2.0/-18.4)} & \cellcolor{red!13}$\downarrow$1.0 \\
 & 0.7B & -16.6 {\tiny (-1.5/-15.1)} & \cellcolor{red!16}$\downarrow$2.0 & -21.6 {\tiny (-2.4/-19.2)} & \cellcolor{red!13}$\downarrow$1.0 & -11.8 {\tiny (-3.6/-8.3)} & \cellcolor{green!17}$\uparrow$1.0 & -23.5 {\tiny (-2.9/-20.6)} & \cellcolor{red!19}$\downarrow$3.0 & -14.4 {\tiny (-3.1/-11.3)} & \cellcolor{red!13}$\downarrow$1.0 & -6.6 {\tiny (-1.1/-5.5)} & \cellcolor{green!24}$\uparrow$2.0 \\
 & 1.2B & -5.7 {\tiny (-0.3/-5.5)} & \cellcolor{green!24}$\uparrow$2.0 & -4.6 {\tiny (-0.4/-4.2)} & \cellcolor{green!24}$\uparrow$2.0 & -4.9 {\tiny (-0.7/-4.2)} & \cellcolor{green!31}$\uparrow$3.0 & -10.1 {\tiny (-0.4/-9.8)} & \cellcolor{red!13}$\downarrow$1.0 & -13.5 {\tiny (-0.9/-12.6)} & \cellcolor{red!16}$\downarrow$2.0 & -9.9 {\tiny (-0.0/-9.9)} & \cellcolor{red!16}$\downarrow$2.0 \\
 & 2.6B & -0.1 {\tiny (-0.0/-0.1)} & \cellcolor{green!49}$\uparrow$5.5 & -0.2 {\tiny (-0.0/-0.2)} & \cellcolor{green!56}$\uparrow$6.5 & -5.4 {\tiny (-1.1/-4.2)} & \cellcolor{green!35}$\uparrow$3.5 & -1.3 {\tiny (-0.3/-1.0)} & \cellcolor{green!52}$\uparrow$6.0 & -4.5 {\tiny (-1.7/-2.8)} & \cellcolor{green!31}$\uparrow$3.0 & -2.1 {\tiny (-0.1/-2.0)} & \cellcolor{green!17}$\uparrow$1.0 \\
\midrule
\multirow{4}{*}{Llama-3} 
 & 1B & -1.9 {\tiny (-2.4/+0.5)} & \cellcolor{green!45}$\uparrow$5.0 & -1.9 {\tiny (-2.1/+0.2)} & \cellcolor{green!45}$\uparrow$5.0 & +2.0 {\tiny (-3.8/+5.8)} & \cellcolor{green!77}$\uparrow$9.5 & -1.2 {\tiny (-2.5/+1.4)} & \cellcolor{green!45}$\uparrow$5.0 & -1.7 {\tiny (-3.2/+1.5)} & \cellcolor{green!38}$\uparrow$4.0 & -1.0 {\tiny (-1.6/+0.6)} & 0.0 \\
 & 3B & -14.8 {\tiny (-6.5/-8.3)} & \cellcolor{red!13}$\downarrow$1.0 & -13.0 {\tiny (-4.3/-8.8)} & \cellcolor{red!13}$\downarrow$1.0 & -2.2 {\tiny (-5.8/+3.6)} & \cellcolor{green!66}$\uparrow$8.0 & -17.9 {\tiny (-11.7/-6.2)} & \cellcolor{red!22}$\downarrow$4.0 & -14.7 {\tiny (-14.0/-0.8)} & \cellcolor{red!16}$\downarrow$2.0 & -15.9 {\tiny (-8.7/-7.2)} & \cellcolor{red!19}$\downarrow$3.0 \\
 & 8B & -38.1 {\tiny (-0.0/-38.1)} & \cellcolor{red!43}$\downarrow$11.0 & -32.5 {\tiny (-0.2/-32.4)} & \cellcolor{red!42}$\downarrow$10.5 & -5.6 {\tiny (-2.0/-3.6)} & \cellcolor{green!38}$\uparrow$4.0 & -26.4 {\tiny (-0.5/-25.9)} & \cellcolor{red!37}$\downarrow$9.0 & -14.3 {\tiny (-2.0/-12.3)} & \cellcolor{red!22}$\downarrow$4.0 & -22.8 {\tiny (-1.0/-21.8)} & \cellcolor{red!34}$\downarrow$8.0 \\
 & 70B & +0.2 {\tiny (-0.0/+0.2)} & \cellcolor{green!31}$\uparrow$3.0 & +0.1 {\tiny (-0.0/+0.1)} & 0.0 & -3.3 {\tiny (-2.7/-0.7)} & 0.0 & -0.6 {\tiny (-0.7/+0.1)} & 0.0 & -3.5 {\tiny (-2.9/-0.6)} & \cellcolor{green!31}$\uparrow$3.0 & +0.4 {\tiny (-0.0/+0.4)} & \cellcolor{green!24}$\uparrow$2.0 \\
\midrule
\multirow{3}{*}{Ministral} 
 & 3B & -2.0 {\tiny (-0.0/-2.0)} & \cellcolor{green!31}$\uparrow$3.0 & -0.7 {\tiny (-0.0/-0.7)} & \cellcolor{green!38}$\uparrow$4.0 & -11.4 {\tiny (-6.7/-4.7)} & \cellcolor{green!14}$\uparrow$0.5 & -2.1 {\tiny (-0.3/-1.7)} & \cellcolor{green!24}$\uparrow$2.0 & -8.0 {\tiny (-5.1/-2.9)} & \cellcolor{green!31}$\uparrow$3.0 & -1.5 {\tiny (-0.0/-1.5)} & \cellcolor{green!14}$\uparrow$0.5 \\
 & 8B & -4.4 {\tiny (-0.0/-4.4)} & 0.0 & -2.6 {\tiny (-0.0/-2.6)} & \cellcolor{green!24}$\uparrow$2.0 & -8.7 {\tiny (-5.6/-3.1)} & \cellcolor{green!24}$\uparrow$2.0 & -8.3 {\tiny (-0.6/-7.7)} & \cellcolor{green!17}$\uparrow$1.0 & -12.3 {\tiny (-3.1/-9.2)} & \cellcolor{green!17}$\uparrow$1.0 & -9.1 {\tiny (-0.0/-9.1)} & \cellcolor{red!25}$\downarrow$5.0 \\
 & 14B & -1.8 {\tiny (-0.0/-1.8)} & 0.0 & -0.9 {\tiny (-0.1/-0.8)} & 0.0 & -10.3 {\tiny (-3.8/-6.5)} & \cellcolor{red!13}$\downarrow$1.0 & -5.7 {\tiny (-0.5/-5.1)} & 0.0 & -11.3 {\tiny (-3.0/-8.3)} & \cellcolor{green!17}$\uparrow$1.0 & -8.1 {\tiny (-0.0/-8.1)} & \cellcolor{red!16}$\downarrow$2.0 \\
\midrule
\multirow{2}{*}{OLMo-3} 
 & 7B & -0.3 {\tiny (-0.2/-0.1)} & \cellcolor{green!45}$\uparrow$5.0 & -0.0 {\tiny (-0.0/-0.0)} & \cellcolor{green!59}$\uparrow$7.0 & -8.0 {\tiny (-6.7/-1.3)} & \cellcolor{green!21}$\uparrow$1.5 & -0.9 {\tiny (-0.5/-0.4)} & \cellcolor{green!52}$\uparrow$6.0 & -8.4 {\tiny (-5.8/-2.6)} & \cellcolor{green!17}$\uparrow$1.0 & +0.4 {\tiny (-0.0/+0.4)} & \cellcolor{green!31}$\uparrow$3.0 \\
 & 32B & +0.2 {\tiny (-0.1/+0.3)} & \cellcolor{green!31}$\uparrow$3.0 & 0.0 {\tiny (-0.0/0.0)} & \cellcolor{green!38}$\uparrow$4.0 & -25.9 {\tiny (-22.3/-3.6)} & \cellcolor{red!43}$\downarrow$11.0 & -1.6 {\tiny (-0.6/-1.1)} & \cellcolor{green!38}$\uparrow$4.0 & -13.0 {\tiny (-9.5/-3.5)} & \cellcolor{red!16}$\downarrow$2.0 & 0.0 {\tiny (-0.0/0.0)} & 0.0 \\
\midrule
\multirow{2}{*}{Phi-4} 
 & 3.6B & -0.4 {\tiny (-0.0/-0.4)} & \cellcolor{green!45}$\uparrow$5.0 & -0.7 {\tiny (-0.0/-0.7)} & \cellcolor{green!38}$\uparrow$4.0 & +3.1 {\tiny (-0.2/+3.3)} & \cellcolor{green!100}$\uparrow$14.0 & -3.0 {\tiny (-0.1/-2.9)} & \cellcolor{green!38}$\uparrow$4.0 & -1.9 {\tiny (-0.4/-1.6)} & \cellcolor{green!38}$\uparrow$4.0 & -0.1 {\tiny (-0.0/-0.1)} & \cellcolor{green!21}$\uparrow$1.5 \\
 & 14B & -6.7 {\tiny (-0.0/-6.7)} & \cellcolor{red!33}$\downarrow$7.5 & -8.0 {\tiny (-0.0/-8.0)} & \cellcolor{red!49}$\downarrow$13.0 & -10.5 {\tiny (-0.2/-10.3)} & \cellcolor{red!13}$\downarrow$1.0 & -12.3 {\tiny (-0.0/-12.3)} & \cellcolor{red!22}$\downarrow$4.0 & -6.5 {\tiny (-0.1/-6.4)} & \cellcolor{red!13}$\downarrow$1.0 & -10.0 {\tiny (-0.0/-10.0)} & \cellcolor{red!25}$\downarrow$5.0 \\
\midrule
\multirow{5}{*}{Qwen-3} 
 & 0.6B & -31.2 {\tiny (-0.0/-31.2)} & \cellcolor{red!16}$\downarrow$2.0 & -43.3 {\tiny (-0.0/-43.3)} & \cellcolor{red!16}$\downarrow$2.0 & -14.7 {\tiny (-0.0/-14.7)} & \cellcolor{red!43}$\downarrow$11.0 & -19.3 {\tiny (-0.0/-19.3)} & 0.0 & -8.0 {\tiny (-0.0/-8.0)} & 0.0 & -8.4 {\tiny (-0.0/-8.4)} & \cellcolor{red!16}$\downarrow$2.0 \\
 & 4B & -0.2 {\tiny (-0.0/-0.2)} & \cellcolor{green!38}$\uparrow$4.0 & +0.1 {\tiny (-0.0/+0.1)} & \cellcolor{green!45}$\uparrow$5.0 & -7.1 {\tiny (-2.0/-5.1)} & \cellcolor{green!38}$\uparrow$4.0 & -3.1 {\tiny (-0.1/-3.0)} & \cellcolor{green!31}$\uparrow$3.0 & -8.5 {\tiny (-0.9/-7.6)} & \cellcolor{green!24}$\uparrow$2.0 & -0.1 {\tiny (-0.0/-0.1)} & \cellcolor{green!31}$\uparrow$3.0 \\
 & 8B & -2.2 {\tiny (-0.0/-2.2)} & \cellcolor{green!31}$\uparrow$3.0 & -0.9 {\tiny (-0.0/-0.9)} & \cellcolor{green!24}$\uparrow$2.0 & -11.6 {\tiny (-2.0/-9.6)} & \cellcolor{red!16}$\downarrow$2.0 & -7.8 {\tiny (-0.3/-7.5)} & 0.0 & -15.1 {\tiny (-1.5/-13.7)} & \cellcolor{red!13}$\downarrow$1.0 & -4.0 {\tiny (-0.0/-4.0)} & \cellcolor{green!24}$\uparrow$2.0 \\
 & 14B & -16.0 {\tiny (-0.0/-16.0)} & \cellcolor{red!64}$\downarrow$18.0 & -11.2 {\tiny (-0.0/-11.2)} & \cellcolor{red!64}$\downarrow$18.0 & -26.1 {\tiny (-2.9/-23.2)} & \cellcolor{red!55}$\downarrow$15.0 & -21.6 {\tiny (-0.0/-21.5)} & \cellcolor{red!52}$\downarrow$14.0 & -30.4 {\tiny (-1.4/-29.0)} & \cellcolor{red!40}$\downarrow$10.0 & -10.3 {\tiny (-0.0/-10.3)} & \cellcolor{red!22}$\downarrow$4.0 \\
 & 32B & -19.5 {\tiny (-0.0/-19.5)} & \cellcolor{red!72}$\downarrow$20.5 & -13.0 {\tiny (-0.0/-13.0)} & \cellcolor{red!70}$\downarrow$20.0 & -38.6 {\tiny (-1.3/-37.3)} & \cellcolor{red!88}$\downarrow$26.0 & -27.9 {\tiny (-0.0/-27.9)} & \cellcolor{red!67}$\downarrow$19.0 & -37.7 {\tiny (-1.3/-36.4)} & \cellcolor{red!67}$\downarrow$19.0 & -5.8 {\tiny (-0.0/-5.8)} & \cellcolor{red!13}$\downarrow$1.0 \\
\midrule
\multirow{4}{*}{SmolLM-2/3} 
 & 0.135B & -23.2 {\tiny (-22.9/-0.3)} & \cellcolor{green!17}$\uparrow$1.0 & -24.0 {\tiny (-23.7/-0.3)} & \cellcolor{green!28}$\uparrow$2.5 & -20.8 {\tiny (-20.5/-0.2)} & \cellcolor{red!28}$\downarrow$6.0 & -22.7 {\tiny (-22.0/-0.7)} & 0.0 & -11.9 {\tiny (-11.6/-0.3)} & \cellcolor{red!13}$\downarrow$1.0 & -18.2 {\tiny (-17.9/-0.4)} & \cellcolor{green!17}$\uparrow$1.0 \\
 & 0.36B & -23.9 {\tiny (-1.7/-22.2)} & \cellcolor{green!24}$\uparrow$2.0 & -26.5 {\tiny (-1.3/-25.2)} & \cellcolor{green!21}$\uparrow$1.5 & -21.2 {\tiny (-10.0/-11.2)} & \cellcolor{red!28}$\downarrow$6.0 & -24.1 {\tiny (-2.8/-21.4)} & \cellcolor{green!24}$\uparrow$2.0 & -9.2 {\tiny (-4.0/-5.2)} & \cellcolor{green!31}$\uparrow$3.0 & -22.3 {\tiny (-0.9/-21.4)} & 0.0 \\
 & 1.7B & -57.2 {\tiny (-0.0/-57.2)} & \cellcolor{red!31}$\downarrow$7.0 & -79.2 {\tiny (-0.0/-79.2)} & \cellcolor{red!37}$\downarrow$9.0 & -27.7 {\tiny (-0.2/-27.5)} & \cellcolor{red!57}$\downarrow$15.5 & -44.9 {\tiny (-0.1/-44.8)} & \cellcolor{red!31}$\downarrow$7.0 & -15.9 {\tiny (-0.1/-15.8)} & \cellcolor{red!13}$\downarrow$1.0 & -18.4 {\tiny (-0.1/-18.2)} & 0.0 \\
 & 3B & -0.3 {\tiny (-0.1/-0.3)} & \cellcolor{green!42}$\uparrow$4.5 & -0.4 {\tiny (-0.1/-0.3)} & \cellcolor{green!52}$\uparrow$6.0 & -6.9 {\tiny (-3.3/-3.6)} & \cellcolor{green!14}$\uparrow$0.5 & -3.0 {\tiny (-1.7/-1.3)} & \cellcolor{green!31}$\uparrow$3.0 & -4.9 {\tiny (-3.1/-1.8)} & \cellcolor{green!24}$\uparrow$2.0 & -1.7 {\tiny (-0.4/-1.3)} & \cellcolor{green!17}$\uparrow$1.0 \\
\bottomrule
\end{tabular}
}
\caption{Effect of regex-based evaluation on performance measurement for multiple-choice benchmarks.}
\label{tab:detailed_results_regex_delta_mc}
\end{table}

\begin{table}[h]
\centering
\small
\resizebox{\textwidth}{!}{
\begin{tabular}{llcccccccccc}
\toprule
\multirow{2}{*}{\textbf{Family}} & \multirow{2}{*}{\textbf{Size}} & \multicolumn{2}{c}{\textbf{AIME24}} & \multicolumn{2}{c}{\textbf{AIME25}} & \multicolumn{2}{c}{\textbf{ASDiv}} & \multicolumn{2}{c}{\textbf{GSM8K}} & \multicolumn{2}{c}{\textbf{Math}} \\
\cmidrule(lr){3-4} \cmidrule(lr){5-6} \cmidrule(lr){7-8} \cmidrule(lr){9-10} \cmidrule(lr){11-12}
 & & \multicolumn{1}{c}{$\Delta$Accuracy} & \multicolumn{1}{c}{$\Delta$Rank} & \multicolumn{1}{c}{$\Delta$Accuracy} & \multicolumn{1}{c}{$\Delta$Rank} & \multicolumn{1}{c}{$\Delta$Accuracy} & \multicolumn{1}{c}{$\Delta$Rank} & \multicolumn{1}{c}{$\Delta$Accuracy} & \multicolumn{1}{c}{$\Delta$Rank} & \multicolumn{1}{c}{$\Delta$Accuracy} & \multicolumn{1}{c}{$\Delta$Rank} \\
\midrule
\multirow{2}{*}{Apertus} 
 & 8B & 0.0 {\tiny (-0.0/0.0)} & \cellcolor{green!30}$\uparrow$4.0 & 0.0 {\tiny (-0.0/0.0)} & \cellcolor{green!35}$\uparrow$5.0 & -5.1 {\tiny (-0.1/-5.0)} & \cellcolor{green!25}$\uparrow$3.0 & -0.4 {\tiny (-0.2/-0.2)} & \cellcolor{green!25}$\uparrow$3.0 & -10.2 {\tiny (-1.2/-9.0)} & \cellcolor{green!40}$\uparrow$6.0 \\
 & 70B & 0.0 {\tiny (-0.0/0.0)} & \cellcolor{green!30}$\uparrow$4.0 & 0.0 {\tiny (-0.0/0.0)} & \cellcolor{green!35}$\uparrow$5.0 & -5.8 {\tiny (-0.7/-5.1)} & \cellcolor{green!15}$\uparrow$1.0 & -0.3 {\tiny (-0.0/-0.3)} & \cellcolor{green!20}$\uparrow$2.0 & -13.3 {\tiny (-1.3/-12.0)} & \cellcolor{green!45}$\uparrow$7.0 \\
\midrule
\multirow{3}{*}{EuroLLM} 
 & 1.7B & 0.0 {\tiny (-0.0/0.0)} & \cellcolor{green!30}$\uparrow$4.0 & 0.0 {\tiny (-0.0/0.0)} & \cellcolor{green!35}$\uparrow$5.0 & -1.7 {\tiny (-0.9/-0.8)} & \cellcolor{green!15}$\uparrow$1.0 & -0.8 {\tiny (-0.2/-0.7)} & \cellcolor{green!15}$\uparrow$1.0 & -2.7 {\tiny (-2.0/-0.7)} & 0.0 \\
 & 9B & -10.0 {\tiny (-6.7/-3.3)} & \cellcolor{red!47}$\downarrow$7.5 & -3.3 {\tiny (-3.3/0.0)} & \cellcolor{red!25}$\downarrow$3.0 & -29.5 {\tiny (-11.9/-17.6)} & \cellcolor{red!30}$\downarrow$4.0 & -27.5 {\tiny (-10.7/-16.8)} & \cellcolor{red!40}$\downarrow$6.0 & -30.2 {\tiny (-15.5/-14.7)} & \cellcolor{red!15}$\downarrow$1.0 \\
 & 22B & -20.0 {\tiny (-3.3/-16.7)} & \cellcolor{red!25}$\downarrow$3.0 & -6.7 {\tiny (-0.0/-6.7)} & \cellcolor{red!47}$\downarrow$7.5 & -32.3 {\tiny (-1.9/-30.4)} & \cellcolor{red!50}$\downarrow$8.0 & -29.7 {\tiny (-5.4/-24.3)} & \cellcolor{red!50}$\downarrow$8.0 & -38.6 {\tiny (-4.2/-34.4)} & \cellcolor{red!47}$\downarrow$7.5 \\
\midrule
\multirow{3}{*}{Falcon-3} 
 & 1B & -3.3 {\tiny (-3.3/0.0)} & \cellcolor{red!15}$\downarrow$1.0 & -3.3 {\tiny (-0.0/-3.3)} & \cellcolor{red!25}$\downarrow$3.0 & -51.1 {\tiny (-45.0/-6.1)} & \cellcolor{red!30}$\downarrow$4.0 & -36.7 {\tiny (-31.1/-5.6)} & \cellcolor{red!30}$\downarrow$4.0 & -22.6 {\tiny (-18.8/-3.8)} & \cellcolor{red!15}$\downarrow$1.0 \\
 & 3B & -6.7 {\tiny (-0.0/-6.7)} & \cellcolor{red!30}$\downarrow$4.0 & -6.7 {\tiny (-0.0/-6.7)} & \cellcolor{red!47}$\downarrow$7.5 & -6.7 {\tiny (-0.6/-6.1)} & 0.0 & -1.9 {\tiny (-0.2/-1.7)} & \cellcolor{green!15}$\uparrow$1.0 & -24.7 {\tiny (-4.1/-20.5)} & \cellcolor{green!15}$\uparrow$1.0 \\
 & 7B & -6.7 {\tiny (-0.0/-6.7)} & \cellcolor{green!32}$\uparrow$4.5 & -6.7 {\tiny (-0.0/-6.7)} & \cellcolor{green!27}$\uparrow$3.5 & -5.8 {\tiny (-0.0/-5.8)} & \cellcolor{green!35}$\uparrow$5.0 & -0.2 {\tiny (-0.1/-0.1)} & \cellcolor{green!20}$\uparrow$2.0 & -24.9 {\tiny (-0.8/-24.1)} & \cellcolor{green!25}$\uparrow$3.0 \\
\midrule
\multirow{4}{*}{Gemma-3} 
 & 1B & -6.7 {\tiny (-6.7/0.0)} & \cellcolor{red!30}$\downarrow$4.0 & -3.3 {\tiny (-3.3/0.0)} & \cellcolor{red!25}$\downarrow$3.0 & -10.8 {\tiny (-5.2/-5.7)} & \cellcolor{red!15}$\downarrow$1.0 & -6.7 {\tiny (-5.5/-1.3)} & \cellcolor{green!20}$\uparrow$2.0 & -28.7 {\tiny (-13.3/-15.5)} & 0.0 \\
 & 4B & -6.7 {\tiny (-0.0/-6.7)} & \cellcolor{green!27}$\uparrow$3.5 & -16.7 {\tiny (-3.3/-13.3)} & \cellcolor{red!27}$\downarrow$3.5 & -7.2 {\tiny (-0.0/-7.1)} & \cellcolor{red!32}$\downarrow$4.5 & -1.3 {\tiny (-0.3/-1.0)} & \cellcolor{green!15}$\uparrow$1.0 & -27.3 {\tiny (-2.5/-24.8)} & \cellcolor{green!25}$\uparrow$3.0 \\
 & 12B & -6.7 {\tiny (-3.3/-3.3)} & \cellcolor{green!55}$\uparrow$9.0 & -6.7 {\tiny (-0.0/-6.7)} & \cellcolor{green!37}$\uparrow$5.5 & -7.2 {\tiny (-0.0/-7.2)} & \cellcolor{red!42}$\downarrow$6.5 & -1.1 {\tiny (-0.1/-1.0)} & \cellcolor{green!20}$\uparrow$2.0 & -29.9 {\tiny (-3.6/-26.3)} & \cellcolor{red!20}$\downarrow$2.0 \\
 & 27B & -10.0 {\tiny (-3.3/-6.7)} & \cellcolor{green!42}$\uparrow$6.5 & -6.7 {\tiny (-0.0/-6.7)} & \cellcolor{green!32}$\uparrow$4.5 & -7.5 {\tiny (-0.1/-7.4)} & \cellcolor{red!60}$\downarrow$10.0 & -0.8 {\tiny (-0.2/-0.6)} & 0.0 & -29.0 {\tiny (-2.0/-27.0)} & \cellcolor{green!15}$\uparrow$1.0 \\
\midrule
\multirow{4}{*}{LFM-2} 
 & 0.35B & 0.0 {\tiny (-0.0/0.0)} & \cellcolor{green!30}$\uparrow$4.0 & 0.0 {\tiny (-0.0/0.0)} & \cellcolor{green!35}$\uparrow$5.0 & -19.5 {\tiny (-1.7/-17.7)} & 0.0 & -22.8 {\tiny (-1.2/-21.6)} & \cellcolor{red!15}$\downarrow$1.0 & -24.0 {\tiny (-14.0/-10.0)} & \cellcolor{red!15}$\downarrow$1.0 \\
 & 0.7B & 0.0 {\tiny (-0.0/0.0)} & \cellcolor{green!30}$\uparrow$4.0 & 0.0 {\tiny (-0.0/0.0)} & \cellcolor{green!35}$\uparrow$5.0 & -5.3 {\tiny (-0.8/-4.5)} & \cellcolor{green!25}$\uparrow$3.0 & -1.4 {\tiny (-1.1/-0.3)} & \cellcolor{green!15}$\uparrow$1.0 & -16.8 {\tiny (-7.5/-9.3)} & \cellcolor{green!25}$\uparrow$3.0 \\
 & 1.2B & -10.0 {\tiny (-3.3/-6.7)} & \cellcolor{red!47}$\downarrow$7.5 & -3.3 {\tiny (-0.0/-3.3)} & \cellcolor{red!25}$\downarrow$3.0 & -8.5 {\tiny (-3.6/-4.9)} & \cellcolor{green!15}$\uparrow$1.0 & -7.1 {\tiny (-6.7/-0.3)} & \cellcolor{green!15}$\uparrow$1.0 & -20.6 {\tiny (-5.0/-15.6)} & \cellcolor{green!25}$\uparrow$3.0 \\
 & 2.6B & -13.3 {\tiny (-3.3/-10.0)} & \cellcolor{green!35}$\uparrow$5.0 & -13.3 {\tiny (-10.0/-3.3)} & \cellcolor{red!25}$\downarrow$3.0 & -4.7 {\tiny (-0.0/-4.6)} & \cellcolor{green!70}$\uparrow$12.0 & -0.6 {\tiny (-0.1/-0.5)} & \cellcolor{green!15}$\uparrow$1.0 & -30.6 {\tiny (-6.9/-23.7)} & \cellcolor{green!20}$\uparrow$2.0 \\
\midrule
\multirow{4}{*}{Llama-3} 
 & 1B & -3.3 {\tiny (-3.3/0.0)} & \cellcolor{red!15}$\downarrow$1.0 & -3.3 {\tiny (-3.3/0.0)} & \cellcolor{red!25}$\downarrow$3.0 & -15.8 {\tiny (-14.5/-1.3)} & 0.0 & -8.7 {\tiny (-8.7/-0.0)} & \cellcolor{green!20}$\uparrow$2.0 & -28.1 {\tiny (-25.5/-2.7)} & \cellcolor{red!20}$\downarrow$2.0 \\
 & 3B & -6.7 {\tiny (-6.7/0.0)} & \cellcolor{green!32}$\uparrow$4.5 & -3.3 {\tiny (-0.0/-3.3)} & \cellcolor{red!25}$\downarrow$3.0 & -4.7 {\tiny (-1.2/-3.5)} & \cellcolor{green!15}$\uparrow$1.0 & -0.5 {\tiny (-0.4/-0.1)} & \cellcolor{green!25}$\uparrow$3.0 & -19.6 {\tiny (-4.5/-15.1)} & \cellcolor{green!25}$\uparrow$3.0 \\
 & 8B & 0.0 {\tiny (-0.0/0.0)} & \cellcolor{green!57}$\uparrow$9.5 & 0.0 {\tiny (-0.0/0.0)} & \cellcolor{green!35}$\uparrow$5.0 & -6.7 {\tiny (-1.6/-5.1)} & \cellcolor{green!20}$\uparrow$2.0 & -0.5 {\tiny (-0.2/-0.2)} & \cellcolor{green!25}$\uparrow$3.0 & -19.8 {\tiny (-2.4/-17.5)} & \cellcolor{green!17}$\uparrow$1.5 \\
 & 70B & -33.3 {\tiny (-33.3/0.0)} & \cellcolor{red!95}$\downarrow$17.0 & -10.0 {\tiny (-10.0/0.0)} & \cellcolor{red!57}$\downarrow$9.5 & -11.4 {\tiny (-5.9/-5.5)} & \cellcolor{red!40}$\downarrow$6.0 & -24.1 {\tiny (-24.2/+0.1)} & \cellcolor{red!100}$\downarrow$18.0 & -73.1 {\tiny (-71.5/-1.6)} & \cellcolor{red!90}$\downarrow$16.0 \\
\midrule
\multirow{3}{*}{Ministral} 
 & 3B & -36.7 {\tiny (-36.7/-0.0)} & \cellcolor{red!67}$\downarrow$11.5 & -20.0 {\tiny (-20.0/0.0)} & \cellcolor{red!35}$\downarrow$5.0 & -6.3 {\tiny (-0.4/-5.9)} & \cellcolor{green!15}$\uparrow$1.0 & -0.6 {\tiny (-0.3/-0.3)} & \cellcolor{green!20}$\uparrow$2.0 & -34.9 {\tiny (-14.4/-20.5)} & \cellcolor{red!25}$\downarrow$3.0 \\
 & 8B & -40.0 {\tiny (-40.0/0.0)} & \cellcolor{red!55}$\downarrow$9.0 & -36.7 {\tiny (-36.7/-0.0)} & \cellcolor{red!57}$\downarrow$9.5 & -6.1 {\tiny (-0.4/-5.7)} & \cellcolor{green!35}$\uparrow$5.0 & -1.9 {\tiny (-0.9/-1.0)} & 0.0 & -33.7 {\tiny (-10.7/-23.0)} & \cellcolor{red!25}$\downarrow$3.0 \\
 & 14B & -50.0 {\tiny (-50.0/0.0)} & \cellcolor{red!92}$\downarrow$16.5 & -33.3 {\tiny (-33.3/-0.0)} & \cellcolor{red!52}$\downarrow$8.5 & -5.9 {\tiny (-0.2/-5.7)} & \cellcolor{green!25}$\uparrow$3.0 & -1.1 {\tiny (-0.7/-0.5)} & \cellcolor{green!12}$\uparrow$0.5 & -37.9 {\tiny (-15.9/-22.0)} & \cellcolor{red!42}$\downarrow$6.5 \\
\midrule
\multirow{2}{*}{OLMo-3} 
 & 7B & -30.0 {\tiny (-20.0/-10.0)} & \cellcolor{red!37}$\downarrow$5.5 & -13.3 {\tiny (-13.3/0.0)} & \cellcolor{green!25}$\uparrow$3.0 & -6.4 {\tiny (-0.1/-6.3)} & \cellcolor{green!12}$\uparrow$0.5 & -0.7 {\tiny (-0.2/-0.5)} & \cellcolor{green!15}$\uparrow$1.0 & -30.3 {\tiny (-5.5/-24.8)} & \cellcolor{red!12}$\downarrow$0.5 \\
 & 32B & -36.7 {\tiny (-36.7/-0.0)} & \cellcolor{red!40}$\downarrow$6.0 & -23.3 {\tiny (-23.3/-0.0)} & \cellcolor{red!20}$\downarrow$2.0 & -5.9 {\tiny (-0.0/-5.9)} & \cellcolor{green!20}$\uparrow$2.0 & -0.8 {\tiny (-0.5/-0.3)} & \cellcolor{red!15}$\downarrow$1.0 & -31.7 {\tiny (-9.5/-22.1)} & \cellcolor{red!25}$\downarrow$3.0 \\
\midrule
\multirow{2}{*}{Phi-4} 
 & 3.6B & -6.7 {\tiny (-6.7/0.0)} & \cellcolor{red!30}$\downarrow$4.0 & -3.3 {\tiny (-3.3/-0.0)} & \cellcolor{green!22}$\uparrow$2.5 & -5.6 {\tiny (-0.0/-5.6)} & \cellcolor{green!40}$\uparrow$6.0 & -1.5 {\tiny (-0.0/-1.5)} & 0.0 & -28.7 {\tiny (-9.8/-18.8)} & \cellcolor{green!15}$\uparrow$1.0 \\
 & 14B & -16.7 {\tiny (-10.0/-6.7)} & \cellcolor{green!32}$\uparrow$4.5 & -6.7 {\tiny (-6.7/0.0)} & \cellcolor{green!27}$\uparrow$3.5 & -6.7 {\tiny (-0.0/-6.7)} & \cellcolor{red!27}$\downarrow$3.5 & -1.3 {\tiny (-0.0/-1.3)} & \cellcolor{green!15}$\uparrow$1.0 & -28.3 {\tiny (-1.5/-26.8)} & \cellcolor{green!20}$\uparrow$2.0 \\
\midrule
\multirow{5}{*}{Qwen-3} 
 & 0.6B & -10.0 {\tiny (-10.0/0.0)} & \cellcolor{green!12}$\uparrow$0.5 & -3.3 {\tiny (-3.3/0.0)} & \cellcolor{red!25}$\downarrow$3.0 & -4.3 {\tiny (-0.1/-4.2)} & \cellcolor{green!25}$\uparrow$3.0 & -3.4 {\tiny (-0.5/-3.0)} & \cellcolor{green!15}$\uparrow$1.0 & -33.0 {\tiny (-15.4/-17.7)} & \cellcolor{red!30}$\downarrow$4.0 \\
 & 4B & -13.3 {\tiny (-13.3/-0.0)} & \cellcolor{green!27}$\uparrow$3.5 & -10.0 {\tiny (-10.0/0.0)} & \cellcolor{green!32}$\uparrow$4.5 & -7.3 {\tiny (-0.1/-7.2)} & \cellcolor{red!35}$\downarrow$5.0 & -1.1 {\tiny (-0.4/-0.7)} & 0.0 & -28.4 {\tiny (-2.8/-25.6)} & \cellcolor{green!20}$\uparrow$2.0 \\
 & 8B & -16.7 {\tiny (-16.7/0.0)} & \cellcolor{green!27}$\uparrow$3.5 & -23.3 {\tiny (-20.0/-3.3)} & \cellcolor{red!35}$\downarrow$5.0 & -7.0 {\tiny (-0.2/-6.8)} & \cellcolor{red!25}$\downarrow$3.0 & -1.4 {\tiny (-0.0/-1.4)} & \cellcolor{green!17}$\uparrow$1.5 & -27.7 {\tiny (-2.7/-25.0)} & \cellcolor{green!20}$\uparrow$2.0 \\
 & 14B & -20.0 {\tiny (-16.7/-3.3)} & \cellcolor{red!15}$\downarrow$1.0 & -20.0 {\tiny (-20.0/0.0)} & \cellcolor{red!15}$\downarrow$1.0 & -6.5 {\tiny (-0.0/-6.5)} & \cellcolor{green!15}$\uparrow$1.0 & -0.2 {\tiny (-0.0/-0.2)} & \cellcolor{green!25}$\uparrow$3.0 & -28.3 {\tiny (-2.2/-26.1)} & \cellcolor{green!20}$\uparrow$2.0 \\
 & 32B & -10.0 {\tiny (-6.7/-3.3)} & \cellcolor{green!37}$\uparrow$5.5 & -6.7 {\tiny (-3.3/-3.3)} & \cellcolor{green!37}$\uparrow$5.5 & -5.8 {\tiny (-0.0/-5.8)} & 0.0 & -0.8 {\tiny (-0.4/-0.5)} & \cellcolor{red!15}$\downarrow$1.0 & -28.3 {\tiny (-2.5/-25.8)} & \cellcolor{green!30}$\uparrow$4.0 \\
\midrule
\multirow{4}{*}{SmolLM-2/3} 
 & 0.135B & 0.0 {\tiny (-0.0/0.0)} & \cellcolor{green!30}$\uparrow$4.0 & 0.0 {\tiny (-0.0/0.0)} & \cellcolor{green!35}$\uparrow$5.0 & -3.1 {\tiny (-3.0/-0.1)} & 0.0 & -0.1 {\tiny (-0.6/+0.5)} & 0.0 & -1.7 {\tiny (-1.7/-0.0)} & 0.0 \\
 & 0.36B & 0.0 {\tiny (-0.0/0.0)} & \cellcolor{green!30}$\uparrow$4.0 & 0.0 {\tiny (-0.0/0.0)} & \cellcolor{green!35}$\uparrow$5.0 & -3.3 {\tiny (-2.4/-1.0)} & \cellcolor{green!15}$\uparrow$1.0 & -1.7 {\tiny (-1.1/-0.6)} & 0.0 & -4.0 {\tiny (-3.2/-0.8)} & \cellcolor{green!15}$\uparrow$1.0 \\
 & 1.7B & 0.0 {\tiny (-0.0/0.0)} & \cellcolor{green!60}$\uparrow$10.0 & 0.0 {\tiny (-0.0/0.0)} & \cellcolor{green!35}$\uparrow$5.0 & -4.7 {\tiny (-1.8/-2.9)} & \cellcolor{green!20}$\uparrow$2.0 & -2.4 {\tiny (-1.7/-0.6)} & \cellcolor{green!20}$\uparrow$2.0 & -14.2 {\tiny (-7.9/-6.3)} & \cellcolor{green!25}$\uparrow$3.0 \\
 & 3B & -10.0 {\tiny (-3.3/-6.7)} & \cellcolor{green!12}$\uparrow$0.5 & -3.3 {\tiny (-3.3/0.0)} & \cellcolor{green!37}$\uparrow$5.5 & -5.7 {\tiny (-0.3/-5.4)} & \cellcolor{green!20}$\uparrow$2.0 & -0.4 {\tiny (-0.2/-0.2)} & \cellcolor{green!20}$\uparrow$2.0 & -33.2 {\tiny (-12.6/-20.6)} & 0.0 \\
\bottomrule
\end{tabular}
}
\caption{Effect of regex-based evaluation on performance measurement for open-form math benchmarks.}
\label{tab:detailed_results_regex_delta_ofm}
\end{table}

\clearpage

\subsection{BERT-as-a-Judge vs. Regex}

In this section, we provide a detailed, unaggregated comparison of regex-based evaluation and BERT-as-a-Judge, reported by model and task. The results are based on the default BERT-as-a-Judge configuration, trained on 1M question-candidate-reference triplets, with candidates generated under the soft-constraint instruction (\autoref{sec:prompting_details}).

\begin{table}[h]
\centering
\small
\begin{tabular}{llcccccc}
\toprule
\multirow{2}{*}{\textbf{Family}} & \multirow{2}{*}{\textbf{Size}} & \textbf{ARC} & \textbf{ARC} & \multirow{2}{*}{\textbf{GPQA}} & \multirow{2}{*}{\textbf{MMLU}} & \textbf{MMLU} & \textbf{Truthful} \\
 &  & \textbf{Challenge} & \textbf{Easy} &  &  & \textbf{Pro} & \textbf{QA} \\
\midrule
\multirow{2}{*}{Apertus} & 8B & 99.4 & 99.6 & 92.4 & 98.2 & 96.7 & 99.1 \\
 & 70B & 99.5 & 99.7 & 95.1 & 98.4 & 96.6 & 99.0 \\
\midrule
\multirow{3}{*}{EuroLLM} & 1.7B & 99.5 & 99.7 & 92.4 & 97.0 & 88.5 & 83.0 \\
 & 9B & 99.9 & 100.0 & 94.9 & 99.3 & 97.5 & 99.8 \\
 & 22B & 99.7 & 99.9 & 94.4 & 98.4 & 97.2 & 99.9 \\
\midrule
\multirow{3}{*}{Falcon-3} & 1B & 98.4 & 98.9 & 92.4 & 97.9 & 95.0 & 97.6 \\
 & 3B & 99.7 & 99.7 & 94.6 & 99.2 & 97.1 & 99.8 \\
 & 7B & 99.9 & 99.9 & 96.2 & 99.5 & 97.7 & 99.6 \\
\midrule
\multirow{4}{*}{Gemma-3} & 1B & 99.4 & 99.8 & 90.6 & 97.5 & 96.5 & 99.5 \\
 & 4B & 99.8 & 99.9 & 96.0 & 98.3 & 97.0 & 99.8 \\
 & 12B & 99.6 & 99.9 & 94.6 & 99.1 & 97.6 & 99.4 \\
 & 27B & 99.7 & 99.8 & 95.5 & 98.9 & 98.0 & 99.8 \\
\midrule
\multirow{4}{*}{LFM-2} & 0.35B & 98.8 & 98.6 & 89.3 & 95.2 & 93.2 & 93.5 \\
 & 0.7B & 99.1 & 99.2 & 91.5 & 96.8 & 96.1 & 98.2 \\
 & 1.2B & 98.7 & 99.4 & 94.4 & 97.5 & 96.6 & 98.2 \\
 & 2.6B & 99.5 & 99.7 & 94.9 & 98.5 & 96.1 & 99.0 \\
\midrule
\multirow{4}{*}{Llama-3} & 1B & 99.1 & 99.5 & 91.5 & 97.7 & 96.5 & 99.1 \\
 & 3B & 99.5 & 99.7 & 93.5 & 98.0 & 96.5 & 99.1 \\
 & 8B & 99.9 & 99.8 & 94.0 & 98.9 & 97.2 & 99.0 \\
 & 70B & 99.7 & 99.8 & 94.4 & 99.0 & 97.6 & 98.4 \\
\midrule
\multirow{3}{*}{Ministral-3} & 3B & 99.5 & 99.8 & 90.4 & 98.7 & 94.0 & 99.3 \\
 & 8B & 99.7 & 99.8 & 93.5 & 99.1 & 96.3 & 99.3 \\
 & 14B & 99.7 & 99.9 & 94.2 & 99.2 & 97.2 & 99.0 \\
\midrule
\multirow{2}{*}{OLMo-3} & 7B & 99.4 & 99.8 & 92.6 & 98.7 & 94.1 & 99.6 \\
 & 32B & 99.5 & 99.9 & 77.7 & 99.2 & 91.4 & 99.8 \\
\midrule
\multirow{2}{*}{Phi-4} & 3.6B & 99.7 & 99.9 & 94.2 & 99.4 & 97.8 & 99.6 \\
 & 14B & 99.9 & 100.0 & 96.0 & 99.2 & 98.0 & 99.0 \\
\midrule
\multirow{5}{*}{Qwen-3} & 0.6B & 99.8 & 99.9 & 95.8 & 99.1 & 97.4 & 100.0 \\
 & 4B & 99.4 & 99.7 & 95.5 & 98.8 & 97.5 & 99.6 \\
 & 8B & 99.0 & 99.5 & 94.2 & 98.8 & 97.2 & 99.4 \\
 & 14B & 99.7 & 99.9 & 94.0 & 99.2 & 97.8 & 99.6 \\
 & 32B & 99.4 & 99.7 & 96.7 & 99.1 & 97.9 & 99.5 \\
\midrule
\multirow{4}{*}{SmolLM-2/3} & 0.135B & 97.6 & 98.9 & 88.6 & 96.7 & 97.0 & 96.7 \\
 & 0.36B & 99.5 & 99.4 & 97.3 & 98.7 & 98.1 & 98.7 \\
 & 1.7B & 99.9 & 99.9 & 97.5 & 99.5 & 98.7 & 99.9 \\
 & 3B & 99.5 & 99.7 & 94.4 & 98.6 & 97.0 & 98.8 \\
\bottomrule
\end{tabular}
\caption{BERT-as-a-Judge assessment accuracy on multiple-choice benchmarks.}
\label{tab:detailed_results_encoder_mc}
\end{table}

\begin{table}[h]
\centering
\small
\begin{tabular}{llcccccc}
\toprule
\multirow{2}{*}{\textbf{Family}} & \multirow{2}{*}{\textbf{Size}} & \textbf{ARC} & \textbf{ARC} & \multirow{2}{*}{\textbf{GPQA}} & \multirow{2}{*}{\textbf{MMLU}} & \textbf{MMLU} & \textbf{Truthful} \\
 &  & \textbf{Challenge} & \textbf{Easy} &  &  & \textbf{Pro} & \textbf{QA} \\
\midrule
\multirow{2}{*}{Apertus} & 8B & 99.0 & 99.2 & 91.5 & 96.4 & 96.2 & 97.8 \\
 & 70B & 97.7 & 97.9 & 93.8 & 94.9 & 94.5 & 96.2 \\
\midrule
\multirow{3}{*}{EuroLLM} & 1.7B & 70.2 & 70.8 & 71.7 & 72.6 & 77.6 & 71.0 \\
 & 9B & 99.5 & 99.9 & 89.7 & 98.0 & 91.3 & 99.8 \\
 & 22B & 91.2 & 92.1 & 85.3 & 85.8 & 87.3 & 93.5 \\
\midrule
\multirow{3}{*}{Falcon-3} & 1B & 73.3 & 66.8 & 84.8 & 84.3 & 85.8 & 93.1 \\
 & 3B & 96.6 & 97.7 & 93.3 & 95.5 & 91.1 & 98.7 \\
 & 7B & 94.6 & 93.3 & 90.4 & 92.8 & 91.8 & 97.8 \\
\midrule
\multirow{4}{*}{Gemma-3} & 1B & 98.7 & 98.9 & 86.4 & 95.3 & 94.1 & 99.5 \\
 & 4B & 99.8 & 99.9 & 95.8 & 98.2 & 96.9 & 99.8 \\
 & 12B & 99.6 & 99.9 & 94.2 & 99.1 & 97.6 & 99.4 \\
 & 27B & 99.7 & 99.8 & 95.8 & 98.9 & 98.1 & 99.8 \\
\midrule
\multirow{4}{*}{LFM-2} & 0.35B & 63.9 & 49.8 & 87.9 & 69.6 & 84.7 & 79.4 \\
 & 0.7B & 82.5 & 78.0 & 86.4 & 75.4 & 84.8 & 92.4 \\
 & 1.2B & 92.1 & 94.2 & 89.7 & 87.5 & 85.1 & 87.6 \\
 & 2.6B & 99.2 & 99.4 & 91.1 & 97.5 & 94.5 & 96.9 \\
\midrule
\multirow{4}{*}{Llama-3} & 1B & 96.9 & 97.6 & 90.4 & 95.9 & 94.7 & 97.8 \\
 & 3B & 84.2 & 86.4 & 89.7 & 79.9 & 83.4 & 83.8 \\
 & 8B & 61.8 & 67.3 & 90.4 & 72.7 & 84.3 & 76.7 \\
 & 70B & 99.7 & 99.8 & 92.6 & 98.0 & 95.6 & 97.4 \\
\midrule
\multirow{3}{*}{Ministral-3} & 3B & 98.0 & 99.1 & 88.6 & 97.3 & 91.8 & 97.3 \\
 & 8B & 95.5 & 97.3 & 89.5 & 91.5 & 87.5 & 90.5 \\
 & 14B & 98.2 & 98.9 & 89.3 & 94.2 & 88.6 & 90.5 \\
\midrule
\multirow{2}{*}{OLMo-3} & 7B & 99.2 & 99.8 & 91.1 & 98.2 & 91.4 & 99.6 \\
 & 32B & 99.5 & 99.9 & 74.1 & 98.0 & 86.9 & 99.5 \\
\midrule
\multirow{2}{*}{Phi-4} & 3.6B & 99.2 & 99.1 & 94.2 & 96.2 & 95.9 & 99.1 \\
 & 14B & 93.3 & 92.0 & 86.4 & 87.0 & 92.4 & 88.5 \\
\midrule
\multirow{5}{*}{Qwen-3} & 0.6B & 68.8 & 56.7 & 83.9 & 80.0 & 89.7 & 91.6 \\
 & 4B & 98.6 & 99.5 & 90.6 & 95.2 & 90.3 & 99.6 \\
 & 8B & 96.1 & 98.1 & 85.7 & 91.2 & 84.1 & 94.7 \\
 & 14B & 83.8 & 88.7 & 72.5 & 78.1 & 69.3 & 89.2 \\
 & 32B & 80.0 & 86.5 & 60.9 & 71.7 & 62.0 & 93.5 \\
\midrule
\multirow{4}{*}{SmolLM-2/3} & 0.135B & 76.8 & 76.0 & 79.2 & 77.3 & 88.1 & 81.8 \\
 & 0.36B & 76.1 & 73.5 & 78.3 & 75.8 & 90.6 & 77.7 \\
 & 1.7B & 42.8 & 20.8 & 71.4 & 55.1 & 84.1 & 81.6 \\
 & 3B & 99.0 & 99.2 & 88.6 & 95.6 & 93.7 & 96.3 \\
\bottomrule
\end{tabular}
\caption{Regex assessment accuracy on multiple-choice benchmarks.}
\label{tab:detailed_results_regex_mc}
\end{table}

\begin{table}[h]
\centering
\small
\begin{tabular}{llcccc}
\toprule
\textbf{Family} & \textbf{Size} & \textbf{CoQA} & \textbf{DROP} & \textbf{HotpotQA} & \textbf{SQuAD-v2} \\
\midrule
\multirow{2}{*}{Apertus} & 8B & 90.2 & 89.1 & 91.9 & 89.9 \\
 & 70B & 87.4 & 89.8 & 91.8 & 89.0 \\
\midrule
\multirow{3}{*}{EuroLLM} & 1.7B & 91.4 & 91.3 & 92.1 & 91.2 \\
 & 9B & 90.0 & 89.6 & 93.5 & 89.6 \\
 & 22B & 90.4 & 89.2 & 93.3 & 89.4 \\
\midrule
\multirow{3}{*}{Falcon-3} & 1B & 90.4 & 90.2 & 90.4 & 91.5 \\
 & 3B & 88.6 & 89.3 & 89.0 & 89.0 \\
 & 7B & 92.8 & 92.2 & 93.8 & 89.1 \\
\midrule
\multirow{4}{*}{Gemma-3} & 1B & 84.8 & 90.1 & 91.0 & 91.1 \\
 & 4B & 75.4 & 83.5 & 88.9 & 87.8 \\
 & 12B & 90.6 & 85.2 & 91.8 & 85.2 \\
 & 27B & 87.2 & 85.4 & 88.7 & 83.9 \\
\midrule
\multirow{4}{*}{LFM-2} & 0.35B & 88.8 & 90.4 & 90.7 & 92.1 \\
 & 0.7B & 83.4 & 89.0 & 85.2 & 92.1 \\
 & 1.2B & 91.4 & 90.5 & 91.5 & 91.5 \\
 & 2.6B & 85.6 & 85.2 & 88.8 & 88.9 \\
\midrule
\multirow{4}{*}{Llama-3} & 1B & 91.6 & 89.4 & 94.5 & 92.7 \\
 & 3B & 88.4 & 88.8 & 91.1 & 89.7 \\
 & 8B & 88.4 & 87.8 & 92.8 & 88.3 \\
 & 70B & 89.8 & 89.0 & 90.5 & 87.3 \\
\midrule
\multirow{3}{*}{Ministral-3} & 3B & 75.6 & 85.0 & 86.4 & 87.7 \\
 & 8B & 87.2 & 87.1 & 89.9 & 87.5 \\
 & 14B & 88.6 & 89.3 & 91.0 & 87.7 \\
\midrule
\multirow{2}{*}{OLMo-3} & 7B & 92.2 & 85.0 & 92.1 & 88.2 \\
 & 32B & 92.0 & 81.7 & 91.5 & 87.3 \\
\midrule
\multirow{2}{*}{Phi-4} & 3.6B & 90.0 & 90.5 & 90.8 & 89.8 \\
 & 14B & 82.2 & 89.2 & 89.4 & 88.2 \\
\midrule
\multirow{5}{*}{Qwen-3} & 0.6B & 86.0 & 90.5 & 92.9 & 91.5 \\
 & 4B & 85.4 & 88.8 & 91.4 & 90.2 \\
 & 8B & 91.6 & 84.6 & 90.5 & 89.0 \\
 & 14B & 93.0 & 90.6 & 93.5 & 87.3 \\
 & 32B & 91.2 & 89.2 & 92.2 & 88.4 \\
\midrule
\multirow{4}{*}{SmolLM-2/3} & 0.135B & 86.8 & 92.0 & 88.3 & 91.1 \\
 & 0.36B & 85.2 & 89.7 & 90.1 & 91.2 \\
 & 1.7B & 91.8 & 93.5 & 93.8 & 92.0 \\
 & 3B & 84.8 & 88.7 & 88.1 & 89.9 \\
\bottomrule
\end{tabular}
\caption{BERT-as-a-Judge assessment accuracy on context extraction benchmarks.}
\label{tab:detailed_results_encoder_ce}
\end{table}

\begin{table}[h]
\centering
\begin{tabular}{llcccc}
\toprule
\textbf{Family} & \textbf{Size} & \textbf{CoQA} & \textbf{DROP} & \textbf{HotpotQA} & \textbf{SQuAD-v2} \\
\midrule
\multirow{2}{*}{Apertus} & 8B & 67.0 & 75.7 & 73.7 & 78.3 \\
 & 70B & 56.6 & 76.3 & 72.5 & 72.8 \\
\midrule
\multirow{3}{*}{EuroLLM} & 1.7B & 49.2 & 80.0 & 60.5 & 69.0 \\
 & 9B & 70.8 & 79.9 & 80.5 & 76.1 \\
 & 22B & 72.4 & 73.0 & 80.2 & 69.4 \\
\midrule
\multirow{3}{*}{Falcon-3} & 1B & 55.4 & 85.1 & 77.5 & 78.8 \\
 & 3B & 61.6 & 79.2 & 71.3 & 76.1 \\
 & 7B & 77.4 & 83.9 & 83.8 & 80.1 \\
\midrule
\multirow{4}{*}{Gemma-3} & 1B & 75.0 & 88.8 & 83.4 & 82.9 \\
 & 4B & 54.6 & 70.3 & 71.2 & 69.5 \\
 & 12B & 68.8 & 66.9 & 76.0 & 69.0 \\
 & 27B & 68.4 & 67.4 & 69.6 & 68.3 \\
\midrule
\multirow{4}{*}{LFM-2} & 0.35B & 76.0 & 90.2 & 84.0 & 86.0 \\
 & 0.7B & 62.0 & 86.6 & 78.1 & 84.2 \\
 & 1.2B & 80.4 & 86.3 & 82.9 & 84.3 \\
 & 2.6B & 66.4 & 69.7 & 72.2 & 64.7 \\
\midrule
\multirow{4}{*}{Llama-3} & 1B & 66.4 & 75.1 & 74.8 & 79.0 \\
 & 3B & 56.8 & 79.4 & 72.8 & 69.6 \\
 & 8B & 74.0 & 77.1 & 79.9 & 76.5 \\
 & 70B & 78.4 & 81.1 & 82.3 & 70.2 \\
\midrule
\multirow{3}{*}{Ministral-3} & 3B & 62.8 & 65.5 & 66.6 & 67.5 \\
 & 8B & 67.2 & 63.8 & 71.0 & 64.5 \\
 & 14B & 70.6 & 68.2 & 73.0 & 62.3 \\
\midrule
\multirow{2}{*}{OLMo-3} & 7B & 78.4 & 74.8 & 83.3 & 71.8 \\
 & 32B & 72.0 & 70.3 & 78.9 & 62.5 \\
\midrule
\multirow{2}{*}{Phi-4} & 3.6B & 56.8 & 75.0 & 66.5 & 63.4 \\
 & 14B & 63.2 & 66.9 & 68.7 & 53.0 \\
\midrule
\multirow{5}{*}{Qwen-3} & 0.6B & 70.4 & 88.9 & 86.6 & 86.7 \\
 & 4B & 71.4 & 76.8 & 81.0 & 73.2 \\
 & 8B & 67.8 & 62.5 & 75.9 & 64.2 \\
 & 14B & 77.8 & 79.2 & 83.3 & 67.5 \\
 & 32B & 75.4 & 73.0 & 79.1 & 63.2 \\
\midrule
\multirow{4}{*}{SmolLM-2/3} & 0.135B & 60.8 & 87.4 & 67.0 & 75.6 \\
 & 0.36B & 45.0 & 81.8 & 61.4 & 71.3 \\
 & 1.7B & 72.4 & 90.5 & 81.5 & 80.1 \\
 & 3B & 62.8 & 76.9 & 71.5 & 73.0 \\
\bottomrule
\end{tabular}
\caption{Regex assessment accuracy on context extraction benchmarks.}
\label{tab:detailed_results_regex_ce}
\end{table}

\begin{table}[h]
\centering
\small
\begin{tabular}{llccccc}
\toprule
\textbf{Family} & \textbf{Size} & \textbf{AIME24} & \textbf{AIME25} & \textbf{ASDiv} & \textbf{GSM8K} & \textbf{Math} \\
\midrule
\multirow{2}{*}{Apertus} & 8B & 100.0 & 100.0 & 95.4 & 98.6 & 93.3 \\
 & 70B & 100.0 & 100.0 & 94.5 & 99.1 & 92.6 \\
\midrule
\multirow{3}{*}{EuroLLM} & 1.7B & 100.0 & 100.0 & 97.8 & 97.6 & 97.0 \\
 & 9B & 90.0 & 96.7 & 95.4 & 98.6 & 91.8 \\
 & 22B & 80.0 & 93.3 & 95.1 & 98.9 & 90.2 \\
\midrule
\multirow{3}{*}{Falcon-3} & 1B & 96.7 & 96.7 & 95.6 & 97.9 & 90.6 \\
 & 3B & 96.7 & 93.3 & 95.4 & 99.4 & 93.3 \\
 & 7B & 93.3 & 93.3 & 95.2 & 99.8 & 93.9 \\
\midrule
\multirow{4}{*}{Gemma-3} & 1B & 93.3 & 100.0 & 93.5 & 96.5 & 90.4 \\
 & 4B & 90.0 & 83.3 & 95.2 & 98.6 & 93.6 \\
 & 12B & 96.7 & 93.3 & 94.8 & 98.9 & 94.9 \\
 & 27B & 90.0 & 93.3 & 95.1 & 99.2 & 95.4 \\
\midrule
\multirow{4}{*}{LFM-2} & 0.35B & 100.0 & 100.0 & 96.2 & 97.2 & 92.3 \\
 & 0.7B & 100.0 & 100.0 & 94.8 & 98.9 & 93.1 \\
 & 1.2B & 90.0 & 96.7 & 94.6 & 98.9 & 92.9 \\
 & 2.6B & 90.0 & 90.0 & 95.2 & 99.2 & 94.9 \\
\midrule
\multirow{4}{*}{Llama-3} & 1B & 96.7 & 96.7 & 93.7 & 97.6 & 92.9 \\
 & 3B & 96.7 & 96.7 & 94.9 & 99.2 & 93.0 \\
 & 8B & 100.0 & 100.0 & 94.0 & 99.0 & 92.5 \\
 & 70B & 93.3 & 93.3 & 95.3 & 99.6 & 94.5 \\
\midrule
\multirow{3}{*}{Ministral-3} & 3B & 66.7 & 56.7 & 94.4 & 99.2 & 90.3 \\
 & 8B & 66.7 & 63.3 & 94.9 & 99.8 & 92.8 \\
 & 14B & 70.0 & 60.0 & 95.3 & 99.6 & 92.5 \\
\midrule
\multirow{2}{*}{OLMo-3} & 7B & 80.0 & 80.0 & 95.5 & 98.9 & 94.7 \\
 & 32B & 60.0 & 66.7 & 95.6 & 99.3 & 93.6 \\
\midrule
\multirow{2}{*}{Phi-4} & 3.6B & 96.7 & 100.0 & 95.4 & 99.5 & 94.6 \\
 & 14B & 83.3 & 100.0 & 95.2 & 99.7 & 95.9 \\
\midrule
\multirow{5}{*}{Qwen-3} & 0.6B & 86.7 & 100.0 & 95.7 & 97.0 & 92.4 \\
 & 4B & 86.7 & 93.3 & 94.9 & 99.2 & 95.3 \\
 & 8B & 83.3 & 83.3 & 95.0 & 99.4 & 95.9 \\
 & 14B & 80.0 & 80.0 & 95.4 & 99.7 & 95.4 \\
 & 32B & 93.3 & 93.3 & 95.3 & 99.6 & 95.8 \\
\midrule
\multirow{4}{*}{SmolLM-2/3} & 0.135B & 100.0 & 100.0 & 98.3 & 98.6 & 97.7 \\
 & 0.36B & 100.0 & 100.0 & 98.0 & 98.9 & 95.7 \\
 & 1.7B & 100.0 & 100.0 & 95.8 & 98.0 & 93.1 \\
 & 3B & 93.3 & 96.7 & 95.2 & 99.0 & 94.6 \\
\bottomrule
\end{tabular}
\caption{BERT-as-a-Judge assessment accuracy on open-form math benchmarks.}
\label{tab:detailed_results_encoder_ofm}
\end{table}

\begin{table}[h]
\centering
\small
\begin{tabular}{llccccc}
\toprule
\textbf{Family} & \textbf{Size} & \textbf{AIME24} & \textbf{AIME25} & \textbf{ASDiv} & \textbf{GSM8K} & \textbf{Math} \\
\midrule
\multirow{2}{*}{Apertus} & 8B & 100.0 & 100.0 & 94.0 & 99.2 & 88.4 \\
 & 70B & 100.0 & 100.0 & 92.9 & 99.7 & 85.0 \\
\midrule
\multirow{3}{*}{EuroLLM} & 1.7B & 100.0 & 100.0 & 96.4 & 97.5 & 96.5 \\
 & 9B & 90.0 & 96.7 & 70.2 & 72.5 & 69.4 \\
 & 22B & 80.0 & 93.3 & 66.9 & 70.3 & 61.1 \\
\midrule
\multirow{3}{*}{Falcon-3} & 1B & 96.7 & 96.7 & 48.6 & 63.3 & 77.0 \\
 & 3B & 93.3 & 93.3 & 92.5 & 97.8 & 74.7 \\
 & 7B & 93.3 & 93.3 & 92.5 & 99.8 & 74.5 \\
\midrule
\multirow{4}{*}{Gemma-3} & 1B & 93.3 & 96.7 & 88.7 & 93.1 & 71.1 \\
 & 4B & 93.3 & 83.3 & 91.5 & 98.6 & 72.3 \\
 & 12B & 93.3 & 93.3 & 91.9 & 98.9 & 69.8 \\
 & 27B & 90.0 & 93.3 & 91.5 & 99.2 & 70.6 \\
\midrule
\multirow{4}{*}{LFM-2} & 0.35B & 100.0 & 100.0 & 79.8 & 76.6 & 75.9 \\
 & 0.7B & 100.0 & 100.0 & 93.1 & 97.9 & 81.8 \\
 & 1.2B & 90.0 & 96.7 & 90.5 & 92.6 & 78.1 \\
 & 2.6B & 86.7 & 86.7 & 92.8 & 99.4 & 69.2 \\
\midrule
\multirow{4}{*}{Llama-3} & 1B & 96.7 & 96.7 & 83.7 & 90.8 & 71.8 \\
 & 3B & 93.3 & 96.7 & 93.8 & 99.4 & 79.9 \\
 & 8B & 100.0 & 100.0 & 91.9 & 99.4 & 79.9 \\
 & 70B & 66.7 & 90.0 & 87.3 & 75.7 & 26.9 \\
\midrule
\multirow{3}{*}{Ministral-3} & 3B & 63.3 & 80.0 & 91.4 & 99.4 & 64.9 \\
 & 8B & 60.0 & 63.3 & 91.7 & 98.1 & 66.0 \\
 & 14B & 50.0 & 66.7 & 91.8 & 98.9 & 62.0 \\
\midrule
\multirow{2}{*}{OLMo-3} & 7B & 70.0 & 86.7 & 92.2 & 99.3 & 69.3 \\
 & 32B & 63.3 & 76.7 & 92.4 & 99.2 & 68.3 \\
\midrule
\multirow{2}{*}{Phi-4} & 3.6B & 93.3 & 96.7 & 92.2 & 98.2 & 70.7 \\
 & 14B & 83.3 & 93.3 & 92.1 & 98.7 & 71.3 \\
\midrule
\multirow{5}{*}{Qwen-3} & 0.6B & 90.0 & 96.7 & 93.9 & 96.4 & 66.8 \\
 & 4B & 86.7 & 90.0 & 91.3 & 98.9 & 71.3 \\
 & 8B & 83.3 & 76.7 & 91.7 & 98.5 & 71.8 \\
 & 14B & 80.0 & 80.0 & 91.9 & 99.6 & 71.3 \\
 & 32B & 90.0 & 93.3 & 91.8 & 99.2 & 71.3 \\
\midrule
\multirow{4}{*}{SmolLM-2/3} & 0.135B & 100.0 & 100.0 & 95.9 & 98.7 & 97.5 \\
 & 0.36B & 100.0 & 100.0 & 95.5 & 98.0 & 94.6 \\
 & 1.7B & 100.0 & 100.0 & 94.0 & 96.9 & 85.1 \\
 & 3B & 90.0 & 96.7 & 92.6 & 99.6 & 66.3 \\
\bottomrule
\end{tabular}
\caption{Regex assessment accuracy on open-form math benchmarks.}
\label{tab:detailed_results_regex_ofm}
\end{table}

\clearpage

\section{Illustrative Examples}
\label{sec:examples}

In this section, we present examples of common failure cases in regex-based evaluation. \autoref{tab:example_1} illustrates a case where parsing fails despite the model producing a correct answer, while \autoref{tab:example_2} shows a case where parsing succeeds but additional formatting introduced by the model prevents correct assessment against the reference.

\begin{table}[h]
\centering
\small
\resizebox{\textwidth}{!}{
\begin{tabular}{p{0.15\linewidth} p{0.8\linewidth}}
\toprule

\multirow{1}{*}{\textbf{Question}} &
\begin{minipage}[t]{\linewidth}
\begin{verbatim}
Answer the question based on the provided context.

Context:
A psychological identity relates to self-image (one's mental model of 
oneself), self-esteem, and individuality. Consequently, Weinreich gives 
the definition "A person's identity is defined as the totality of one's 
self-construal, in which how one construes oneself in the present expresses 
the continuity between how one construes oneself as one was in the past and 
how one construes oneself as one aspires to be in the future"; this allows 
for definitions of aspects of identity, such as: "One's ethnic identity is 
defined as that part of the totality of one's self-construal made up of 
those dimensions that express the continuity between one's construal of 
past ancestry and one's future aspirations in relation to ethnicity" 
(Weinreich, 1986a). 

The description or representation of individual and group identity is a 
central task for psychologists, sociologists and anthropologists and those 
of other disciplines where "identity" needs to be mapped and defined. How 
should one describe the identity of another, in ways which encompass both 
their idiosyncratic qualities and their group memberships or 
identifications, both of which can shift according to circumstance? 
Following on from the work of Kelly, Erikson, Tajfel and others Weinreich's 
Identity Structure Analysis (ISA), is "a structural representation of the 
individual's existential experience, in which the relationships between 
self and other agents are organised in relatively stable structures over 
time … with the emphasis on the socio-cultural milieu in which self 
relates to other agents and institutions" (Weinreich and Saunderson, (eds) 
2003, p1). Using constructs drawn from the salient discourses of the 
individual, the group and cultural norms, the practical operationalisation 
of ISA provides a methodology that maps how these are used by the 
individual, applied across time and milieus by the "situated self" to 
appraise self and other agents and institutions (for example, resulting in 
the individual's evaluation of self and significant others and 
institutions).[citation needed]

Question: What is the ISA?

Conclude your response with "Final answer: X", where X is the exact span 
from the context that answers the question.
\end{verbatim}
\end{minipage} \\

\addlinespace[2pt]
\midrule

\multirow{1}{*}{\textbf{Reference}} &
\begin{minipage}[t]{\linewidth}
\begin{verbatim}
Identity Structure Analysis (ISA)
\end{verbatim}
\end{minipage} \\

\addlinespace[2pt]
\midrule

\multirow{1}{*}{\textbf{Candidate}} &
\begin{minipage}[t]{\linewidth}
\begin{verbatim}
The ISA, or Identity Structure Analysis, is a methodology developed by 
Weinreich that maps how individuals perceive themselves and others in 
relation to their social environment, including their own group memberships 
and identifications, as well as the cultural norms and discourses that 
shape these perceptions.
\end{verbatim}
\end{minipage} \\

\addlinespace[2pt]
\bottomrule
\end{tabular}
}
\caption{Example 488 from the CoQA dataset, along with the answer from Llama-3 1B}
\label{tab:example_1}
\end{table}

\begin{table}[h]
\centering
\small
\resizebox{\textwidth}{!}{
\begin{tabular}{p{0.15\linewidth} p{0.8\linewidth}}
\toprule

\multirow{1}{*}{\textbf{Question}} &
\begin{minipage}[t]{\linewidth}
\begin{verbatim}
We have a triangle $\triangle ABC$ such that $AB = 6,$ $BC = 8,$ and 
$CA = 10.$ If $AD$ is an angle bisector such that $D$ is on $BC,$ then 
find the value of $AD^2.$

Conclude your response with "Final answer: X", where X is the computed 
solution.
\end{verbatim}
\end{minipage} \\

\addlinespace[2pt]
\midrule

\multirow{1}{*}{\textbf{Reference}} &
\begin{minipage}[t]{\linewidth}
\begin{verbatim}
45
\end{verbatim}
\end{minipage} \\

\addlinespace[2pt]
\midrule

\multirow{1}{*}{\textbf{Candidate}} &
\begin{minipage}[t]{\linewidth}
\begin{verbatim}
To solve for \( AD^2 \) in \(\triangle ABC\) where \( AB = 6 \), \( BC = 8 \), 
and \( CA = 10 \), and \( AD \) is the angle bisector of \(\angle BAC\) with 
\( D \) on \( BC \), we can use the Angle 
Bisector Theorem and the Angle Bisector Length Formula.

### Step-by-Step Solution:

1. **Angle Bisector Theorem:**
The Angle Bisector Theorem states that the angle bisector divides the 
opposite side into segments proportional to the adjacent sides. Therefore, 
if \( D \) divides \( BC \) into segments \( BD \) and \( DC \), then:
\[
\frac{BD}{DC} = \frac{AB}{AC} = \frac{6}{10} = \frac{3}{5}
\]

[...]

4. **Calculate \( AD^2 \):**
\[
AD^2 = 6 \cdot 10 \left( 1 - \frac{8^2}{(6 + 10)^2} \right)
\]
\[
AD^2 = 60 \left( 1 - \frac{64}{16^2} \right)
\]
\[
AD^2 = 60 \left( 1 - \frac{64}{256} \right)
\]
\[
AD^2 = 60 \left( 1 - \frac{1}{4} \right)
\]
\[
AD^2 = 60 \left( \frac{3}{4} \right)
\]
\[
AD^2 = 60 \cdot 0.75
\]
\[
AD^2 = 45
\]

Final answer: \( \boxed{45} \)
\end{verbatim}
\end{minipage} \\

\addlinespace[2pt]
\bottomrule
\end{tabular}
}
\caption{Example 2070 from the MATH dataset, along with the answer from Falcon-3 7B}
\label{tab:example_2}
\end{table}

\end{document}